\documentclass{article}

\PassOptionsToPackage{numbers, compress}{natbib}

\usepackage[preprint]{neurips_2024}
\usepackage[linesnumbered,ruled,vlined,noend]{algorithm2e}
\usepackage{caption}
\usepackage{ulem}
\usepackage{adjustbox}
\usepackage{float}
\usepackage{diagbox}
\usepackage{color}
\usepackage{colortbl}
\usepackage{lipsum}
\usepackage[table]{xcolor}
\definecolor{hiscolor}{RGB}{220,224,228}
\definecolor{applegreen}{rgb}{0.55, 0.71, 0.0}
\definecolor{downred}{RGB}{255, 136, 132}
\definecolor{rolecolor_1}{RGB}{255,247,172}
\definecolor{rolecolor_2}{RGB}{221,241,243}
\definecolor{rolecolor_3}{RGB}{236,244,221}
\definecolor{intercolor_1}{RGB}{255,230,230}
\definecolor{intercolor_2}{RGB}{221,242,247}
\definecolor{bestcolor}{RGB}{142,139,254}
\usepackage{multirow}
\usepackage{enumitem}
\usepackage{booktabs}
\usepackage{array}
\usepackage{makecell}
\usepackage{graphicx}
\usepackage{wrapfig}
\usepackage[utf8]{inputenc}
\usepackage{amsmath}

\definecolor{markcolor}{RGB}{220,224,228}
\usepackage[utf8]{inputenc} 
\usepackage[T1]{fontenc}    
\usepackage{hyperref}       
\usepackage{url}            
\usepackage{booktabs}       
\usepackage{amsfonts}       
\usepackage{nicefrac}       
\usepackage{microtype}      
\usepackage{xcolor}         
\usepackage{soul}

\title{Compressing Long Context for Enhancing RAG with AMR-based Concept Distillation}

\author{Kaize Shi $^{\heartsuit}$, Xueyao Sun$^{\heartsuit,\clubsuit}$, Qing Li $^{\clubsuit}$, Guandong Xu$^{\heartsuit,\diamondsuit}$\\ \\
  $^{\heartsuit}$University of Technology Sydney\\
  $^{\clubsuit}$The Hong Kong Polytechnic University\\
  $^{\diamondsuit}$The Education University of Hong Kong
}

\begin{document}
\maketitle

\begin{abstract}

Large Language Models (LLMs) have made significant strides in information acquisition. However, their overreliance on potentially flawed parametric knowledge leads to hallucinations and inaccuracies, particularly when handling long-tail, domain-specific queries. Retrieval Augmented Generation (RAG) addresses this limitation by incorporating external, non-parametric knowledge. Nevertheless, the retrieved long-context documents often contain noisy, irrelevant information alongside vital knowledge, negatively diluting LLMs' attention. Inspired by the supportive role of essential concepts in individuals' reading comprehension, we propose a novel concept-based RAG framework with the Abstract Meaning Representation (AMR)-based concept distillation algorithm. The proposed algorithm compresses the cluttered raw retrieved documents into a compact set of crucial concepts distilled from the informative nodes of AMR by referring to reliable linguistic features. The concepts explicitly constrain LLMs to focus solely on vital information in the inference process. We conduct extensive experiments on open-domain question-answering datasets to empirically evaluate the proposed method's effectiveness. The results indicate that the concept-based RAG framework outperforms other baseline methods, particularly as the number of supporting documents increases, while also exhibiting robustness across various backbone LLMs. This emphasizes the distilled concepts are informative for augmenting the RAG process by filtering out interference information. To the best of our knowledge, this is the first work introducing AMR to enhance the RAG, presenting a potential solution to augment inference performance with semantic-based context compression.

\end{abstract}

\section{Introduction}

Large Language Models (LLMs) have emerged as indispensable tools for daily information acquisition, owing to their extensive knowledge base and ability to fulfil diverse user instructions~\cite{NEURIPS2020_1457c0d6, touvron2023llama, achiam2023gpt}. By leveraging large-scale pre-training on massive datasets, LLMs memorize vast amounts of knowledge within their parameters as internal memory, known as parametric knowledge~\cite{NEURIPS2022_6f1d43d5}. However, the presence of outdated or incorrect knowledge within internal memory can lead to hallucinations, hindering the performance of LLMs' inferencing process~\cite{tonmoy2024comprehensive}. This limitation is particularly pronounced when handling long-tail knowledge for domain-specific or highly specialized queries, as the inherent difficulty in memorizing rare entities persists even in the most robust models. Consequently, the overreliance on potentially flawed parametric knowledge can significantly interfere with the reliability of LLMs' outputs, especially in scenarios with fine-grained knowledge requirements~\cite{zhang2023siren, muhlgay2023generating}.

Retrieval Augmented Generation (RAG) employs additional retrievers to augment LLMs with external, non-parametric knowledge, effectively expanding their internal knowledge boundaries~\cite{NEURIPS2020_6b493230, gao2023retrieval}. This allows LLMs to access up-to-date, query-focused information that may not be adequately memorized within their parametric memory to alleviate the aforementioned limitations~\cite{karpukhin-etal-2020-dense}. In contrast to fine-tuning by updating the model parameters, RAG preserves pre-trained knowledge while dynamically incorporating relevant external context. This paradigm offers greater flexibility and scalability, as the retrievers can be easily plug-and-play without modifying the underlying language model's parameters, thus circumventing complex computational hurdles~\cite{pmlr-v119-guu20a, gupta2024rag}. However, RAG is easily confused when dealing with long contextual retrieved support documents, which often consist of multiple shreds of evidence for providing vital knowledgeable context but are also accompanied by noisy and irrelevant information~\cite{10.1145/3397271.3401160}. The distracting contexts can dilute the LLMs' attention and adversely affect their performance with misrepresentation~\cite{liu2024lost, krishna-etal-2021-hurdles}. Compressing lengthy contexts to distil vital knowledge is crucial for enhancing LLMs and ensuring factually consistent responses in the RAG process.

\begin{wrapfigure}{r}{0.5\textwidth}
\centering
\includegraphics[width=1\linewidth]{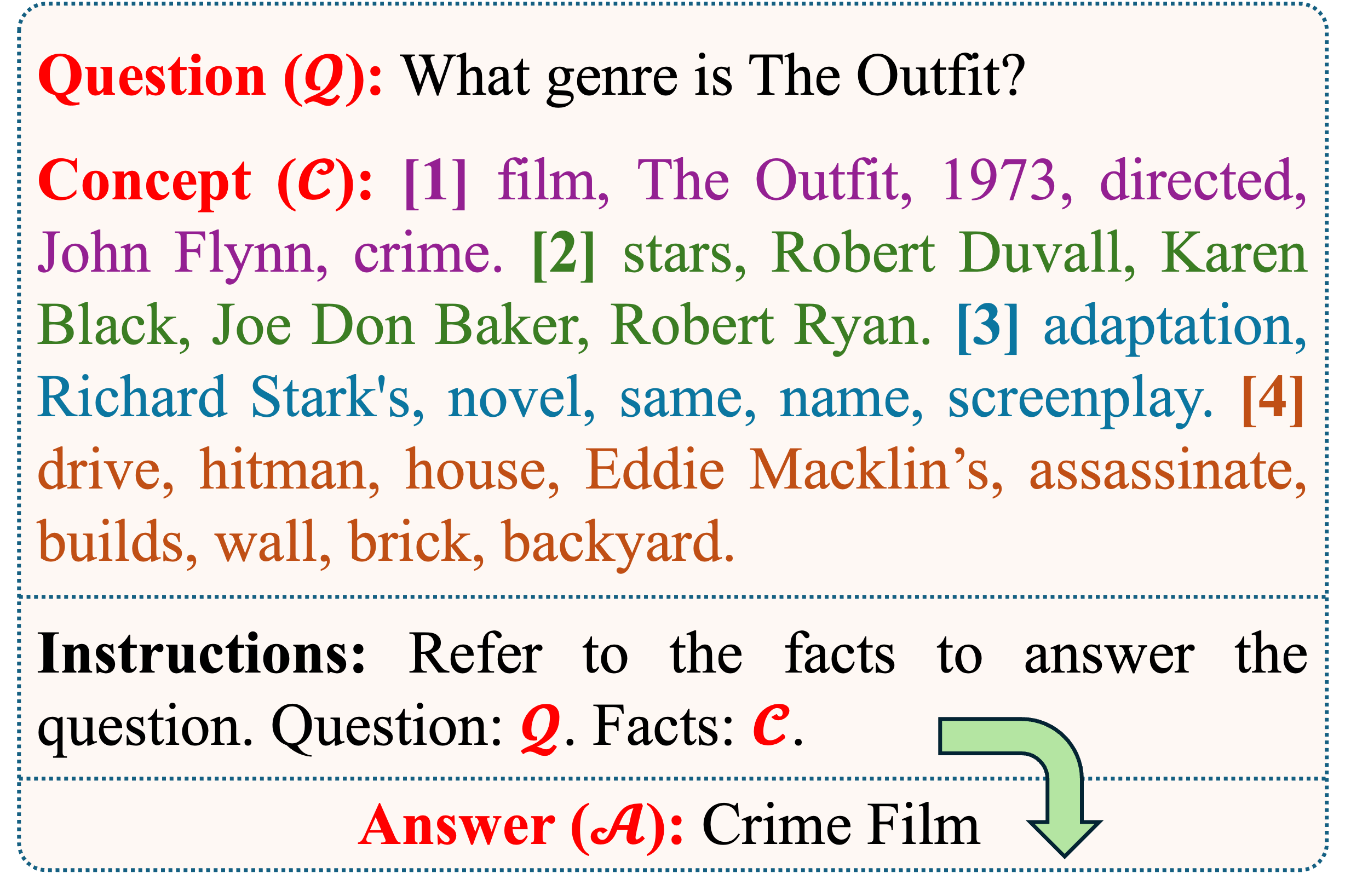}
\caption{The examples of concept-based RAG\protect\footnotemark.}
\label{fig:toy_example}
\end{wrapfigure}
\footnotetext{The corresponding complete sentences: \textcolor[RGB]{148,32,146}{\textbf{[1]} The Outfit is a 1973 crime film directed by John Flynn.} \textcolor[RGB]{59,125,35}{\textbf{[2]} It stars Robert Duvall, Karen Black, Joe Don Baker and Robert Ryan.} \textcolor[RGB]{11,118,160}{\textbf{[3]} Flynn's screenplay is an adaptation of the novel of the same name by Richard Stark.} \textcolor[RGB]{192,79,21}{\textbf{[4]} Two hitmen drive to Eddie Macklin's house to assassinate him as he builds a brick wall in his backyard.}}

Numerous studies have demonstrated that individuals tend to directly search for key concepts when reading long documents as the brain will complete the remaining details based on prior knowledge, expectations, background, and motivations~\cite{graesser1994constructing, johnson2007transposed}. This selective attention to critical information allows ignoring redundant details and rearranging the text informatively~\cite{wit1999linguistic}. As illustrated in Fig.~\ref{fig:toy_example}, given only the key concepts of the question-related supporting documents that still enable us to grasp the crucial semantics. LLMs parameterize massive common knowledge, enabling them to exhibit a similar ability in context understanding even when the word or character-level information is disrupted~\cite{sinha-etal-2021-masked, cao-etal-2023-unnatural}. This provides the possibility of whether LLMs can comprehend scenarios solely based on discrete informative concepts. Linguistic features, such as semantic and syntactic, have significantly improved the interpretability, controllability, and diversity of Natural Language Generation (NLG)~\cite{li-etal-2023-explicit}. Language models can implicitly discover these features during pre-training to ensure the logic of the generated text~\cite{jawahar-etal-2019-bert}. It has been demonstrated that explicitly leveraging linguistic features for downstream tasks is beneficial, as it refactors the source documents into concise representations that reduce entropy by focusing on the critical information, thereby aiding in a comprehensive understanding of the described scenarios~\cite{shannon1948mathematical, wang-etal-2018-tree, sun-etal-2021-aesop, li-etal-2023-explicit, jung2024information, yu2022survey}. This advantage enables the stable linguistic features to reliably assist context understanding.

Inspired by the aforementioned insights, we propose enhancing RAG's performance with the crucial concepts distilled from the raw retrieved supporting documents. To effectively capture the informative concepts, we introduce Abstract Meaning Representation (AMR), a semantic formalism that encodes the meaning of serialized texts by a rooted, directed, labelled, acyclic graph~\cite{banarescu2013abstract}. Compared to other linguistic representations, AMR prioritizes semantic consistency among concepts carried by nodes when representing sentences, offering the advantage of automatically rectifying surface-level variations or understanding abbreviated terms, ensuring the structured concepts represent the underlying meaning to transcend the limitations of linguistic noise~\cite{zhang-etal-2021-fine}. Specifically, we propose the concept-based RAG framework with the AMR-based concept distillation algorithm, which formats the concepts for augmenting LLMs by compressing the lengthy context to concentrate on crucial information exclusively. We empirically experiment on two open-domain Q\&A datasets, PopQA~\cite{mallen-etal-2023-trust} and EntityQuestions~\cite{sciavolino-etal-2021-simple}. The results show that the performance of our method improves significantly as the number of supporting documents increases, outperforming baselines with various compression methods and backbone LLMs. The contributions of this paper can be summarized as follows:

\begin{itemize}
\item This paper proposes the concept-based RAG framework that explicitly integrates AMR, a semantic representation, to enable LLMs to focus on essential rather than messy knowledge when processing long-context retrieved supporting documents. To the best of our knowledge, this is the first research introducing AMR to enhance RAG for more reliable inference.
\item We propose an AMR-based concept distillation algorithm, which compresses long-context raw supporting documents into concepts by formatting the informative nodes. The distilled concepts are more knowledge-centralized than the raw supporting documents, reducing the interference of irrelevant information during the inference process of LLMs.
\item We conduct extensive experiments on open-domain Q\&A datasets. The results indicate that our framework effectively enhances inference performance as the number of supporting documents increases, outperforming baselines with various context compression methods and backbone LLMs. This demonstrates its applicability in long-context RAG scenarios.
\end{itemize}

\section{Related Works}
\label{gen_inst}

\subsection{Long-context Understanding}

The increasing complexity of downstream tasks and the demand for models capable of capturing intricate dependencies have driven significant attention to the long-context understanding of LLMs~\cite{pawar2024and, huang2023advancing, xu2023retrieval}. One prominent research avenue involves modifying the basic architecture of LLMs. For instance, Dai et al.\cite{dai-etal-2019-transformer} introduced a segment-level recurrence mechanism with their Transformer-XL model, enabling it to retain longer contextual information than the standard Transformer structure. Similarly, Beltagy et al.\cite{beltagy2020longformer} extended the self-attention mechanism in their Longformer model to handle longer sequences by introducing a sparse attention pattern, thereby facilitating the efficient processing of documents with thousands of tokens. However, a significant drawback of modifying model architecture is the necessity for complex re-training processes. In contrast, research on prompt compression aims to understand long-token prompts by compressing them into low-dimensional soft prompts~\cite{wingate-etal-2022-prompt, chevalier-etal-2023-adapting, mu2024learning}. While offering a more efficient alternative to architecture modification, this approach constrains the transferability of learned prompts across various LLMs.

Recent research has advanced to a more intuitive level, aiming to comprehensively understand the context by directly expanding the context window or explicit compression. Chen et al.\cite{chen2023extending} introduced position interpolation to extend the context window of pre-trained LLMs, scaling LLaMA's context window to 32k tokens with few fine-tuning steps. Ding et al.\cite{ding2024longrope} proposed LongRoPE to extend LLMs' context window to 2048k tokens while maintaining the performance of the original short context window through a positional and interpolation progressive extension strategy. However, the long context window raises another challenge of diluting core information with redundant data~\cite{xu2023retrieval}. To address this, Li et al.\cite{li-etal-2023-compressing} filtered out irrelevant context with low self-information for compressing the long prompts. Chuang et al.\cite{chuang2024learning} proposed the Nano-Capsulator to compress original prompts into capsule prompts, decreasing inference latency across diverse LLMs. Compression methods can benefit the RAG by allowing LLMs to focus on essential knowledge in supporting documents~\cite{yan2024corrective}.

\subsection{Linguistics-augmented NLG}

Incorporating linguistic principles into LLMs has shown promise in improving the coherence and semantic fidelity of generated text~\cite{yu2022survey}. Augmentation techniques like syntactic trees~\cite{mueller-linzen-2023-plant} and lexical patterns~\cite{li-etal-2023-explicit} assist in linguistic feature injection, enabling language models to generate more faithful text. Ahmed et al.~\cite{Ahmed2024Automatic} proposed automatic semantic augmentation of prompts to enhance LLMs with tagged facts, resulting in improved code summarization performance. Zhou et al.~\cite{pmlr-v202-zhou23g} introduced InstructCTG, a framework for controlling LLMs' generation based on syntax constraints, facilitating flexibility and adaptation to new conditions without complex model modification. LLMs can be explicitly guided by leveraging linguistic insights to mitigate biases inherent in parameterized-only approaches, hereby enhancing performance in tasks demanding strict factual consistency.

Abstract Meaning Representation (AMR) has proven its efficacy in enhancing downstream generation tasks by providing a structured semantic representation that encapsulates static concepts~\cite{NEURIPS2021_479b4864}. Frisoni et al.~\cite{Frisoni_Italiani_Salvatori_Moro_2023} integrated AMR with pre-trained language models to enhance biomedical summarization by capturing inter-entity relations. Ribeiro et al.~\cite{ribeiro-etal-2022-factgraph} employed AMR to improve factuality evaluation in abstractive summarization by identifying content verifiability errors and subsentence-level factual inconsistencies. Shi et al.~\cite{shi-etal-2023-amr} proposed AMR-TST, which generates fluent and reliable texts with the target style by optimizing core concept nodes. Jangra et al.~\cite{jangra-etal-2022-star} preserved style-agnostic content while generating transferred text by utilizing AMR as an intermediate representation. These studies illustrate AMR's advantages in capturing essential concepts containing informative linguistic features.

\section{Method}
\label{headings}

\subsection{Concept-based RAG Framework}

\begin{figure*}[htbp]
 \centering
 \includegraphics[width=1\textwidth]{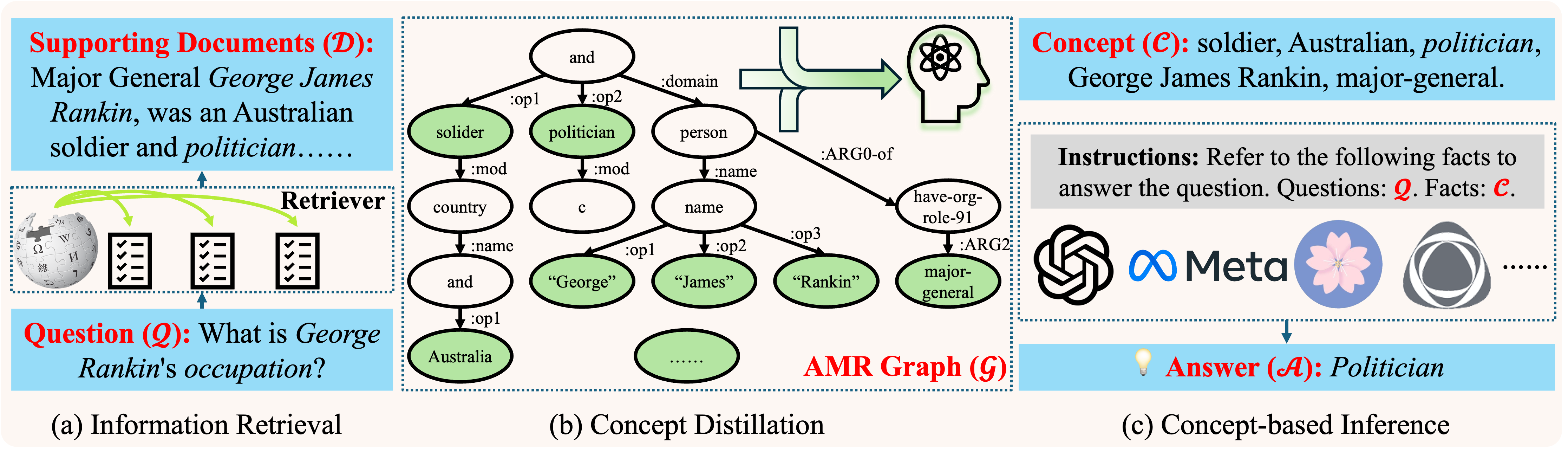}
 \caption{The overview of the concept-based RAG framework, which consists of three main components: (a) information retrieval, (b) concept distillation, and (c) concept-based inference.}
 \label{fig:method_overview}
\end{figure*}

This section introduces the proposed concept-based RAG framework for inference utilising the concepts distilled from the raw supporting documents. The overview of the framework is in Fig.~\ref{fig:method_overview}.

Given an input question $\mathcal{Q}$, the (a) information retrieval component aims to utilize a retriever to return the top-$\mathcal{K}$ knowledgeable supporting documents $\mathcal{D} = \{\mathcal{D}_1, ..., \mathcal{D}_\mathcal{K}\}$ relevant to $\mathcal{Q}$ from sources such as Wikipedia or other information repositories. At this stage, the retriever's performance significantly influences the resulting answer set $\mathcal{A} = \{\mathcal{A}_1, ..., \mathcal{A}_M\}$~\cite{NEURIPS2022_6f1d43d5, gao2023retrieval}. However, the retriever's performance is beyond this paper's scope. We hypothesize that all retrieved supporting documents $\mathcal{D}$ contain the correct answer corresponding to $\mathcal{Q}$, expressed as a proposition: $\forall \mathcal{D}_k \in \mathcal{D}$, $\exists \mathcal{A}_m \in \mathcal{A}$, $\mathcal{A}_m \subseteq \mathcal{D}_k$.

The (b) concept distillation component is devised to format the concept $\mathcal{C}$ from the retrieved supporting document $\mathcal{D}$ by the proposed AMR-based concept distillation algorithm. This algorithm converts the supporting documents from continuous sequences to discrete concepts formatted from the AMR graph, denoted as $\mathcal{G}$. Further details of this algorithm will be elucidated in the subsequent section.

After obtaining the distilled concept $\mathcal{C}$, the (c) concept-based inference component proceeds to integrate it with various backbone LLMs to derive answers $\mathcal{A}$ using a faithful-intensive prompt template as follows: [\textit{Refer to the following facts to answer the question. Facts: $\mathcal{C}$. Question: $\mathcal{Q}$}]. The intensity of prompts has been demonstrated to influence LLMs' adherence to knowledge from internal memory and retrieved documents~\cite{wu2024faithful}. Since our hypothesis is that the retrieved documents contain correct answers, we encourage the LLMs to leverage the knowledge encapsulated in $\mathcal{C}$ when responding to queries. This strategy helps minimize potential conflicts caused by their memorized parametric knowledge. To achieve this objective, we designate the concept as a "\textit{fact}" within the instructional prompt, explicitly delineating a delimited sandbox for LLMs to presuppose the absolute correctness of the knowledge conveyed by $\mathcal{C}$. This non-parametric knowledge can seamlessly integrate into LLMs in a plug-and-play manner. The overarching framework can be represented as Eq.~\ref{eq:framework}.

\vspace{-0.3cm}
\begin{equation}
P(\mathcal{A}|\mathcal{Q}) = P(\mathcal{A}|\mathcal{C},\mathcal{Q})P(\mathcal{C}|\mathcal{D},\mathcal{Q})P(\mathcal{D}|\mathcal{Q}).
\label{eq:framework}
\end{equation}

\vspace{-0.3cm}

\subsection{AMR-based Concept Distillation}

Abstract Meaning Representation (AMR) serves as a logical formal semantic structure proficient in encapsulating common-sense knowledge necessary for representing events, time, participants, and other elements within serialized texts~\cite{santana2023survey}. Given a supporting document $\mathcal{D}_k \in \mathcal{D}$, the AMR parser is utilized to parse $\mathcal{D}_k$ into the corresponding AMR graph $\mathcal{G} = <\mathcal{N},\mathcal{E}>$, where $\mathcal{C}$ represents the nodes for concepts and $\mathcal{E}$ denotes the edges for the correlation relationships. In this context, we utilize a mBart-based~\cite{liu2020multilingual} parser\footnote{\url{https://github.com/BramVanroy/multilingual-text-to-amr}} trained on the AMR 3.0 corpus\footnote{\url{https://catalog.ldc.upenn.edu/LDC2020T02}} to address potential multilingual concerns. The detailed illustration of the AMR graph parsing is depicted in Table~\ref{tab:AMR_tc}.

\begin{wrapfigure}{R}{0.66\textwidth}
\begin{minipage}{0.66\textwidth}
\IncMargin{1em}
\begin{algorithm}[H]
\caption{Concept Distillation}
\label{alg:c_d}
\SetAlgoLined
\SetKwInOut{Input}{Input}
\SetKwInOut{Output}{Output}
\SetKwInOut{Repeat}{repeat}
\Input{AMR Graph ($\mathcal{G}$)}
\Output{concept ($\mathcal{C}$)}
\SetKwFunction{SplitSnt}{SplitSnt}
\SetKwFunction{AppendConcept}{AppendConcept}
\SetKwFunction{SearchRole}{SearchRole}
\SetKwFunction{AppendRole}{AppendRole}
\SetKwFunction{HandleRole}{HandleRole}
\SetKwFunction{IsRole}{IsRole}
\SetKwFunction{ConceptBacktrace}{ConceptBacktrace}
\SetKwFunction{ConceptFormat}{ConceptFormat}
\SetKwFunction{IsWiki}{IsWiki}
\SetKwFunction{HandleWiki}{HandleWiki}
\SetKwFunction{IsDate}{IsDate}
\SetKwFunction{HandleDate}{HandleDate}
\SetKwFunction{IsName}{IsName}
\SetKwFunction{HandleName}{HandleName}
\SetKwFunction{DFS}{DFS}
\SetKwProg{Fn}{Function}{:}{}
\Fn{Concept\_Distillation\textup{($\mathcal{G}$)}}{
    concept $\gets []$, role $\gets []$\;
    \For{$\mathcal{G}_{snt_n}$ in \SplitSnt \textup{($\mathcal{G}$)}}{
        \For{$\mathcal{N}$ in \DFS{$\mathcal{G}_{snt_n}$}}{
            \uIf{\IsRole{$\mathcal{N}$}}{
                \uIf{\IsName{$\mathcal{N}$}}{
                    \AppendRole{\HandleName{$\mathcal{N}$}}
                }
                \uIf{\IsWiki{$\mathcal{N}$}}{
                    \AppendRole{\HandleWiki{$\mathcal{N}$}}
                }
                \uIf{\IsDate{$\mathcal{N}$}}{
                    \AppendRole{\HandleDate{$\mathcal{N}$}}
                }
            }
            \uElse{
                \uIf{\textup{role} is not None}{
                    \AppendConcept{\HandleRole{\textup{role}}}\;
                    role $\gets []$\;
                }
                \AppendConcept{$\mathcal{N}$}\;
            }
            \uIf{\textup{($\mathcal{N}$} is Last\textup{)} and \textup{(role} is not \textup{None)}}{
            \Repeat{Algorithm.Line 5-11}
            \AppendConcept{\HandleRole{\textup{role}}}\;
            }
        }
    }
    concept $\gets$ \ConceptFormat(concept)\;
    concept $\gets$ \ConceptBacktrace(concept)\;
    \Return $\mathcal{C}$ $\gets$ concept
}
\end{algorithm}
\DecMargin{1em}
\end{minipage}
\end{wrapfigure}
We propose the concept distillation algorithm to format the concepts represented in $\mathcal{G}$, as described in Algorithm~\ref{alg:c_d}. The supporting document $\mathcal{D}_k$ encompasses multiple sentences ($snt_n$), and the AMR parser can structurally parse $\mathcal{D}_k$ into a pre-defined \texttt{multi-sentence} structure. The \texttt{SplitSnt($\cdot$)} function is designed to partition $\mathcal{G}$ and organize the resulting sentence-based sub-graphs according to the sequential order. Notably, we simplify $\mathcal{G}$ by disregarding the agent and patient of the concepts, i.e., the edges denoting relations between the connected concepts (Frame args, \texttt{ARGX}). Consequently, $\mathcal{G}$ is streamlined into a unidirectional connecting structure. Leveraging this structure, we perform a Depth First Search, \texttt{DFS($\cdot$)} on the $\mathcal{N}$ of $\mathcal{G}$ to traverse the concepts while maintaining the relative positional correlation of adjacent nodes. This approach emphasizes the connection as it exists in the preceding sequential representation, and the process is elaborated in Fig.~\ref{fig:AMR_DFS}. Previous research has investigated the influence of context order on LLMs~\cite{liu2024lost}. We delve into the various traversal methods for testing their potential impact in Section~\ref{sec:random_order}.

The AMR defines a set of roles to meticulously delineate the semantic fabric of sentences. This paper underscores the meticulous handling of three roles, namely \texttt{:name}, \texttt{:wiki}, and \texttt{date-entity}, employing \texttt{IsRole($\cdot$)} to identify the predefined roles comprehensively. The \texttt{:name} role signifies a property node within the AMR graph, signifying entities such as individuals, organizations, or geographic locations. In instances where the concept expressed by \texttt{:name} spans multiple words, the parsing process of AMR decomposes each word within the \texttt{:name} into predicate roles (\texttt{:op}), thereby dispersing the holistic concept across multiple nodes. During the \texttt{DFS($\cdot$)} traversal process, fragmented nodes can potentially confuse LLMs due to incomplete meaning expressions. To maintain the integrity of concepts carried by \texttt{:name}, we introduce \texttt{HandleName($\cdot$)}, organizing predicates in a stack structure. The \texttt{:wiki} role provides reliable external concept references sourced from Wikipedia. For standardizing concepts' diverse expressions referring to the same named entities, we utilize the \texttt{HandleWiki ($\cdot$)} function, which aligns the concepts with the corresponding definitions in Wikipedia. If the concept in \texttt{:name} differs from \texttt{:wiki}, we designate the concept expressed by this node as \texttt{:wiki} to avoid semantic disambiguation. In addition, there is a \texttt{date-entity} role that depicts temporal concepts. In our algorithm, we specifically manage the roles \texttt{:year}, \texttt{:month}, and \texttt{:day} by \texttt{HandleDate ($\cdot$)}. This function consolidates roles under the same \texttt{date-entity}, forming concepts like "19 04 2024" with numerical months translated into textual representations, "19 April 2024", for clear expression. AMR incorporates special numerical annotations for certain parsing nodes, such as \texttt{work-01}, where the number appended to the word indicates different meanings of the same word in distinct contexts as defined in OntoNotes~\cite{weischedel2013ontonotes}. In the RAG scenario, we provide LLMs with supporting documents comprising a set of concepts. This suggests that concepts are understood in relation to relevant contexts rather than in isolation. Therefore, the proposed concept-based RAG framework depends on the contextual learning capability of LLMs to distinguish between polysemous concepts, instead of relying on intricate semantic references. The nodes belonging to the aforementioned roles are integrated into the preliminary concept set with the \texttt{HandleRole($\cdot$)}, while the \texttt{AppendConcept($\cdot$)} directly integrate the remaining nodes based on the corresponding instances.

The structure of AMR comprises a collection of canonical nodes (\texttt{city-district}, \texttt{market-sector}, etc.) designed to enforce knowledge and prevent hallucination regarding entity types. However, in the concept-based RAG scenario, the inference process isn't directly based on AMR but distilled concepts. The auxiliary semantics embedded within these nodes, which are absent in the source supporting documents, may dilute the essence of the core concept. To address this concern, we employ \texttt{ConceptFormat($\cdot$)} to filter out these nodes to reduce the potential interference. Additionally, frequently occurring concepts are filtered out based on their Inverse Document Frequency (IDF). Furthermore, the selection of representations in AMR is based on the principle of abstraction and generalization rather than the exact lexical items. This representation may mislead the nodes into ignoring variations such as tense, which are informative for concept-based RAG without reference annotations. To mitigate this, we develop the \texttt{ConceptBacktrace($\cdot$)} function to maintain consistency with concepts expressed in the source supporting documents. This function facilitates the backtracking of formatted concepts by incorporating representations from the supporting documents, ensuring they closely adhere to the original semantics without deviation. Subsequently, the backtraced concepts serve as the finalized concepts $\mathcal{C}$, providing conceptual support for LLMs in RAG inference.

\section{Experiments}
\label{sec:experiments}

\subsection{Datasets}
\label{sec:datasets}

We conducted extensive experiments to verify the efficacy of the concept-based RAG framework on open-domain Q\&A datasets: PopQA~\cite{mallen-etal-2023-trust} and EntityQuestions~\cite{sciavolino-etal-2021-simple}. Each dataset includes a label ("\textit{hasanswer}") for every supporting document, indicating whether it contains the answer to the associated question. To ensure a focused evaluation, we screened out the "<$\mathcal{Q}$-$\mathcal{A}$-$\mathcal{D}$>" pairs where \textit{hasanswer=True}. This selection criterion accommodates scenarios where all retrieved documents contribute positively to answering questions, thus mitigating interference from extraneous factors. The experiments involved verifying the LLMs' inference performance with different $\mathcal{K}$, which denotes the amount of supporting documents to $\mathcal{Q}$. For the PopQA dataset, we filtered out questions with subject entities having monthly Wikipedia pageviews ($s_{pop}$) $\geq 500$. This step excludes frequently accessed entities, preserving the dataset focused on long-tail knowledge. This approach serves the dual purpose of preventing data contamination and encouraging LLMs to rely more on retrieved documents than memorized knowledge, mitigating potential knowledge conflicts in the RAG process. The statistical results of the number of the selected pairs with different $\mathcal{K}$ settings are in Table~\ref{tab:datasets}.

\begin{table}[htbp]
\centering
\caption{Statistical results of the number of screened-out <$\mathcal{Q}$-$\mathcal{A}$-$\mathcal{D}$> pairs from the datasets.}
\begin{adjustbox}{width=1\linewidth}
\begin{tabular}{c|c|c|c|c|c|c|c|c|c|c}
\hline
$\mathcal{K}$= & 1 & 2 & 3 & 4 & 5 & 6 & 7 & 8 & 9 & 10 \\ \hline \hline
PopQA~\cite{mallen-etal-2023-trust} & 738 & 1307 & 422 & 262 & 161 & 151 & 108 & 79 & 66 & 70 \\ \hline
EntityQuestions~\cite{sciavolino-etal-2021-simple} & 1671 & 1127 & 670 & 454 & 335 & 264 & 196 & 166 & 163 & 103 \\
\hline
\end{tabular}
\end{adjustbox}
\label{tab:datasets}
\end{table}

\subsection{Baselines}
\label{sec:baselines}

The baseline evaluations encompass two aspects: (1) exploration of diverse backbone LLMs, and (2) experimentation with different context compression methods. Specifically, we consider various mainstream LLMs as backbones, including GPT-Neo-1.3B, GPT-Neo-2.7B~\cite{gpt-neo}, OPT-1.3b, OPT-2.7b~\cite{zhang2022opt}, bloom-560m, bloom-1b1, bloom-1b7, bloom-3b~\cite{le2022bloom}, LLaMA-2-7b-chat-hf, LLaMA-2-13b-chat-hf~\cite{touvron2023llama}. The backbone LLMs coupled with the original supporting documents serve as the Vanilla methods. Regarding the alternative aspect, we explore the three context compression methods: context keywords extraction, context summarization, and Selective Context (SelCon)~\cite{li-etal-2023-compressing}. These methods aim to validate the efficacy of context compression while preserving essential information for inference, emphasizing discrete key features, fluent representation, and non-redundant information.

Inspired by \citet{chuang2024learning}, we employ a novel open-access LLM, LLaMA-2-13b-chat-hf~\cite{touvron2023llama}, for context keyword extraction and summarization. This process involves extracting key phrases or terms from the context and generating a concise summary of the provided content, constrained by prompts of "[\texttt{Generate a short summary of the following content.}]" and "[\texttt{Extract a few keywords from the following content.}]". The detailed prompts are available in Appendix~\ref{sec:prompt}. The SelCon enhances the efficiency of LLMs' inference by identifying and eliminating redundant content from the source context for compression. The reduction ratio of the SelCon compared here is set to $0.5$. These baseline settings effectively demonstrate the comprehensive advantages of the proposed algorithm in capturing informative concepts when compared to various alternative compression techniques, whether generative-based or semantic-based methods.

\subsection{Evaluation Metrics}
\label{sec:evaluation}

We employ two metrics to evaluate the concept-based RAG: accuracy ($Acc.$) and integration ($Intg.$). Accuracy ($Acc.$) is determined by assessing whether any answer $\mathcal{A}$ matches any of the gold answers corresponding to the question $\mathcal{Q}$. The integration metric ($Intg.$) is designed to comprehensively evaluate the performance across various $\mathcal{K}$ of the retrieved supporting documents $\mathcal{D}$. Specifically, the $Intg.$ signifies the area beneath the accuracy curve of each model plotted against the X-axis ($\mathcal{K}$). The calculation of $Intg.$ is as Eq.~\ref{eq:Inte}, where $\mathcal{K} \in [x_s,x_e]$, and $x_s$ and $x_e$ represent the minimum and maximum number of supporting documents respectively. A higher value of $Intg.$ indicates superior overall performance. Given that the proposed framework aims to enhance long-context RAG, we segment the evaluation of $Intg.$ into two distinct intervals: normal interval ($I_n=[1,10], \mathcal{K} \in I_n$) and longer interval ($I_l=[6,10], \mathcal{K} \in I_l$). This division is intended to emphasize the effectiveness of the concept-based RAG framework, particularly in scenarios involving longer contexts.

\vspace{-0.4cm}
\begin{equation}
Intg. = \int_{x_s}^{x_e} Acc(x) \, dx \approx \frac{1}{2} \sum_{i=1}^{{x_e}-{x_s}+1} (x_{i} - x_{i-1}) \left[ Acc(x_{i}) + Acc(x_{i-1}) \right]
\label{eq:Inte}
\end{equation}

\vspace{-0.4cm}

\section{Results and Analysis}

\begin{figure*}[htbp]
 \centering
 \includegraphics[width=1\textwidth]{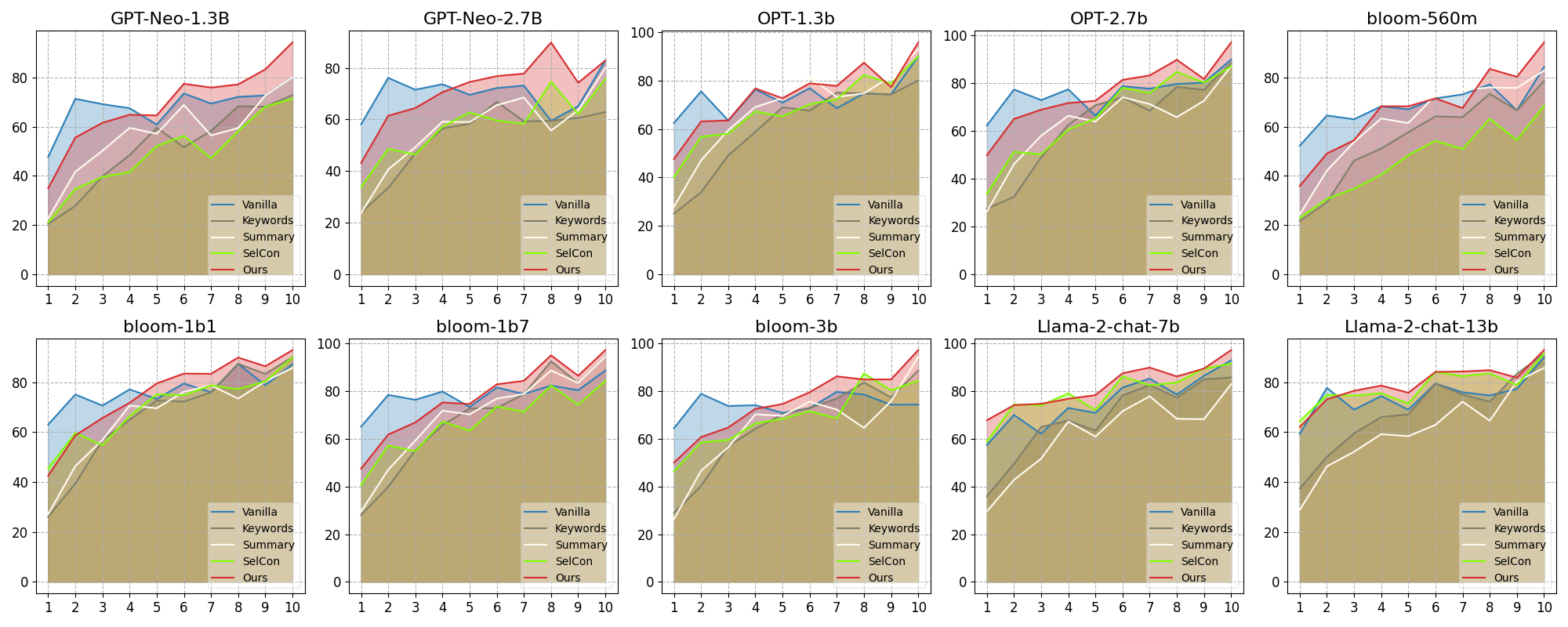}
 \caption{The evaluation results of the $Acc. \uparrow$ trends and $Intg.\uparrow$ on the PopQA dataset. The vertical axis represents $Acc.$, and the horizontal axis represents the number of supporting documents, $\mathcal{K}$. The polyline reflects the changing trend of $Acc.$ with different $\mathcal{K}$, and the under area is $Intg.$}
 \label{fig:PopQA}
\end{figure*}

\begin{figure*}[htbp]
 \centering
 \includegraphics[width=1\textwidth]{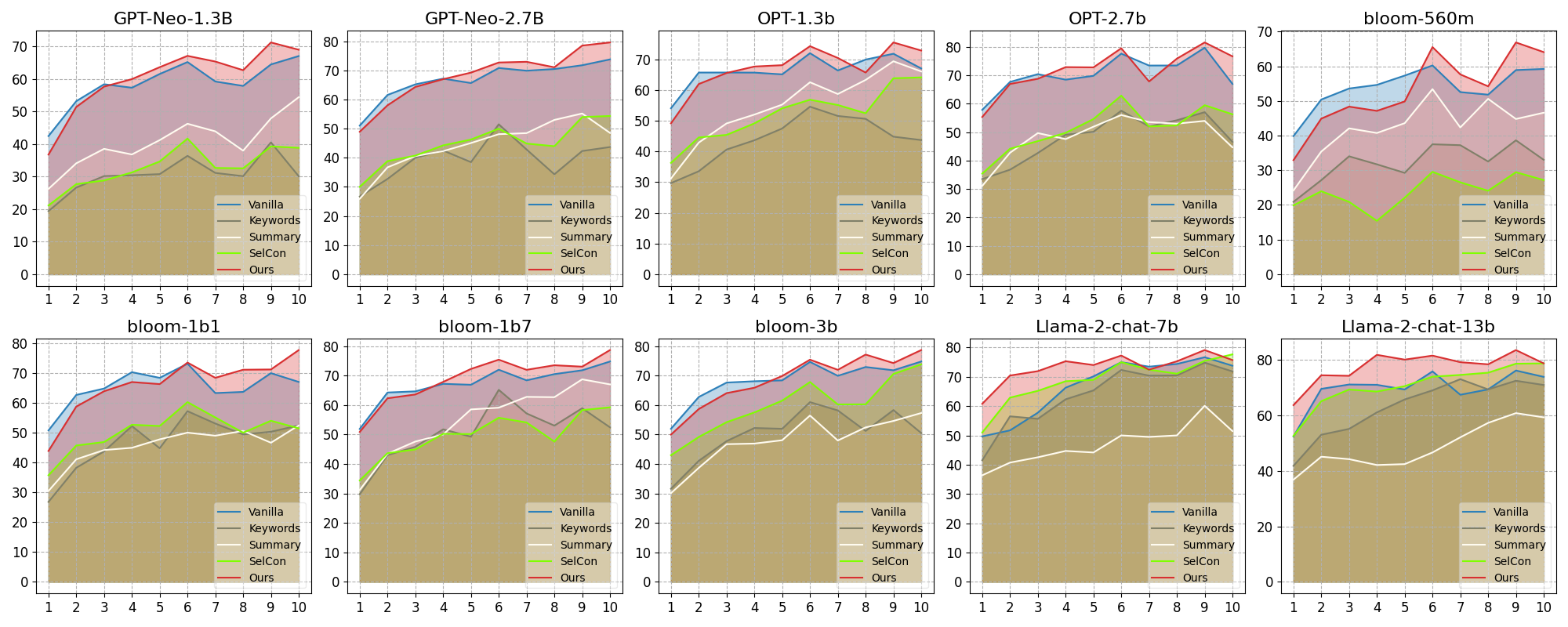}
 \caption{The evaluation results of the $Acc. \uparrow$ trends and $Intg.\uparrow$ on the EntityQuestion dataset. The definitions of the axis and symbols are the same with the Fig.~\ref{fig:PopQA}.}
 \label{fig:EQ}
\end{figure*}

The evaluation results for the PopQA and EntityQuestion datasets are depicted in Fig.~\ref{fig:PopQA} and Fig.~\ref{fig:EQ}, respectively, providing graphical trends of $Acc.$ as $\mathcal{K}$ increases intuitively. Furthermore, Table~\ref{tab:PopQA} and Table~\ref{tab:EQQA} present quantitative results of $Intg.$ for the datasets. These tables include the calculation of $\Delta$, quantifying the improvement achieved by our proposed method over the Vanilla methods. Specifically, $\Delta$ is computed as follows: $\Delta={Intg.}_{ours}-{Intg.}_{vanilla}$. The detailed quantitative evaluation results of $Acc.$ are provided in Table~\ref{tab:aca_PopQA} and Table~\ref{tab:aca_EQQA}. Section~\ref{sec:compression_ratio} and section~\ref{sec:inference_latency} examine compression ratio and inference latency comparison to demonstrate the advantages of concept-compressed contexts.

\begin{table}[htbp]
\centering
\caption{The quantitative results of $Intg.\uparrow$ for the PopQA dataset, where the full name order of the LLMs is: GPT-Neo-1.3B, GPT-Neo-2.7B, OPT-1.3b, OPT-2.7b, bloom-560m, bloom-1b1, bloom-1b7, bloom-3b, LLaMA-2-chat-7b, LLaMA-2-chat-13b. The best results are in \textbf{bold}, and the second best results are in \uline{underlined}. The \textcolor{applegreen}{increased} and \textcolor{downred}{decreased} $\Delta$ are marked differently.}
\begin{adjustbox}{width=1\linewidth}
\begin{tabular}{cccccccccccc}
\hline
$\mathcal{D}$                  & $\mathcal{K} $          & G-1.3 & G-2.7 & O-1.3 & O-2.7 & b-560 & b-1b1 & b-1b7 & b-3 & L-7 & L-13 \\ \hline \hline
\multirow{2}{*}{\rotatebox{0}{\footnotesize{Vanilla}}} & {$I_n$} & \uline{620.68} & \uline{631.39} & \uline{656.68} & \uline{687.15} & \textbf{619.86} & \textbf{692.68} & \textbf{707.25} & \uline{671.88} & 682.30 & 672.03    \\ 
                        
                         & {$I_l$} & \uline{291.08} & \uline{275.32} & 300.85 & 322.23 & 294.94 & 325.37 & 326.29 & 305.91 & 337.19 & 312.62    \\ \hline
\multirow{2}{*}{\rotatebox{0}{\footnotesize{Keywords}}}     & {$I_n$} & 468.94 & 484.98 & 554.67 & 571.38 & 502.70 & 610.69 & 621.85 & 600.65 & 628.78 & 617.06    \\ 
                         
                         & {$I_l$} & 257.12 & 244.24 & 297.70 & 305.64 & 275.39 & \uline{327.70} & \uline{338.01} & \uline{318.37} & 326.41 & 315.93    \\ \hline
\multirow{2}{*}{\rotatebox{0}{\footnotesize{Summary}}}     & {$I_n$} & 517.57 & 513.37 & 619.78 & 575.32 & 573.95 & 608.41 & 637.55 & 591.12 & 564.51 & 553.24    \\ 
                         
                         & {$I_l$} & 263.14 & 260.64 & \uline{316.80} & 290.50 & \uline{304.55} & 313.36 & 336.20 & 297.44 & 291.50 & 291.39    \\ \hline
\multirow{2}{*}{\rotatebox{0}{\footnotesize{SelCon}}}      & {$I_n$} & 444.29 & 524.54 & 615.78 & 607.12 & 423.22 & 634.81 & 606.15 & 625.66 & \uline{715.90} & \uline{703.29}    \\ 
                         
                         & {$I_l$} & 237.49 & 262.78 & 313.39 & \uline{323.69} & 230.20 & 318.64 & 306.72 & 314.07 & \uline{344.10} & \uline{332.51}    \\ \hline
\multirow{2}{*}{\rotatebox{0}{\footnotesize{\textbf{Ours}}}}    & \cellcolor{intercolor_1}{$I_n$} & \cellcolor{intercolor_1}\textbf{625.31} & \cellcolor{intercolor_1}\textbf{652.71} & \cellcolor{intercolor_1}\textbf{668.86} & \cellcolor{intercolor_1}\textbf{688.47} & \cellcolor{intercolor_1}\uline{608.31} & \cellcolor{intercolor_1}\uline{686.29} & \cellcolor{intercolor_1}\uline{698.91} & \cellcolor{intercolor_1}\textbf{681.22} & \cellcolor{intercolor_1}\textbf{738.82} & \cellcolor{intercolor_1}\textbf{716.55} \\ 
                         
                         & \cellcolor{intercolor_2}{$I_l$} & \cellcolor{intercolor_2}\textbf{322.37} & \cellcolor{intercolor_2}\textbf{321.73} & \cellcolor{intercolor_2}\textbf{329.65} & \cellcolor{intercolor_2}\textbf{344.31} & \cellcolor{intercolor_2}\textbf{314.34} & \cellcolor{intercolor_2}\textbf{347.71} & \cellcolor{intercolor_2}\textbf{355.52} & \cellcolor{intercolor_2}\textbf{344.08} & \cellcolor{intercolor_2}\textbf{357.56} & \cellcolor{intercolor_2}\textbf{339.38} \\ \hline \hline
\multirow{2}{*}{\rotatebox{0}{\footnotesize{$\Delta$}}}    & {$I_n$} & \textcolor{applegreen}{+4.63} & \textcolor{applegreen}{+21.32}  & \textcolor{applegreen}{+12.18} & \textcolor{applegreen}{+1.32} & \textcolor{downred}{-11.55} & \textcolor{downred}{-6.93} & \textcolor{downred}{-8.34} & \textcolor{applegreen}{+9.34} & \textcolor{applegreen}{+56.52} & \textcolor{applegreen}{+44.52} \\ 
                         
                         & {$I_l$} & \textcolor{applegreen}{+31.29} & \textcolor{applegreen}{+46.41} & \textcolor{applegreen}{+28.8} & \textcolor{applegreen}{+22.08} & \textcolor{applegreen}{+19.40} & \textcolor{applegreen}{+22.34} & \textcolor{applegreen}{+29.23} & \textcolor{applegreen}{+38.17} & \textcolor{applegreen}{+20.37} & \textcolor{applegreen}{+26.76} \\ \hline
\end{tabular}
\end{adjustbox}
\label{tab:PopQA}
\end{table}

\begin{table}[htbp]
\centering
\caption{The quantitative results of $Intg.\uparrow$ for the EntityQuestions dataset. The LLMs' order and symbol definitions are the same as Table \ref{tab:PopQA}.}
\begin{adjustbox}{width=1\linewidth}
\begin{tabular}{cccccccccccc}
\hline
$\mathcal{D}$                  & $\mathcal{K} $          & G-1.3 & G-2.7 & O-1.3 & O-2.7 & b-560 & b-1b1 & b-1b7 & b-3 & L-7 & L-13 \\ \hline \hline
\multirow{2}{*}{\rotatebox{0}{\footnotesize{Vanilla}}} & {$I_n$} & \uline{531.54} & \uline{605.06} & \uline{602.52} & \uline{634.28} & \textbf{488.95} & \uline{594.88} & \uline{608.85} & \uline{619.30} & 607.22                & 632.24    \\ 
                         & {$I_l$} & \uline{247.50}  & \uline{284.47} & \uline{277.47} & \uline{299.03} & \uline{222.99} & \uline{266.91} & \uline{284.00} & \uline{289.26} & 289.95 & 287.48  \\ \hline
\multirow{2}{*}{\rotatebox{0}{\footnotesize{Keywords}}}     & {$I_n$} & 280.76 & 360.00 & 403.37 & 439.73 & 295.02 & 428.54 & 465.15 & 462.65 & 584.67 & 574.61    \\  
                         & {$I_l$} & 134.96 & 167.13 & 196.04 & 215.41 & 143.68 & 207.59 & 227.84 & 223.38 & 287.84 & 284.53    \\ \hline
\multirow{2}{*}{\rotatebox{0}{\footnotesize{Summary}}}     & {$I_n$} & 366.73 & 406.72 & 501.51 & 446.50 & 388.36 & 415.61 & 501.90 & 435.49 & 425.70 & 438.31    \\
                         & {$I_l$} & 179.97 & 205.02 & 255.51 & 210.93 & 187.75 & 197.43 & 257.16 & 211.83 & 210.34 & 222.92    \\ \hline
\multirow{2}{*}{\rotatebox{0}{\footnotesize{SelCon}}}      & {$I_n$} & 298.49 & 405.22 & 471.36 & 468.18 & 215.52 & 460.37 & 451.41 & 539.49 & \uline{623.91} & \uline{641.01}    \\   
                         & {$I_l$} & 144.69 & 195.05 & 231.76 & 223.55 & 108.45 & 214.94 & 217.40 & 261.79 & \uline{295.33} & \uline{304.57}    \\ \hline
\multirow{2}{*}{\rotatebox{0}{\footnotesize{\textbf{Ours}}}}    & {\cellcolor{intercolor_1}$I_n$} & \cellcolor{intercolor_1}\textbf{551.50} & \cellcolor{intercolor_1}\textbf{618.18} & \cellcolor{intercolor_1}\textbf{609.88} & \cellcolor{intercolor_1}\textbf{652.48} & \cellcolor{intercolor_1}\uline{483.02} & \cellcolor{intercolor_1}\textbf{600.72} & \cellcolor{intercolor_1}\textbf{624.53} & \cellcolor{intercolor_1}\textbf{621.36} & \cellcolor{intercolor_1}\textbf{664.18} & \cellcolor{intercolor_1}\textbf{703.67}    \\   
                         & {\cellcolor{intercolor_2}$I_l$} & \cellcolor{intercolor_2}\textbf{267.12} & \cellcolor{intercolor_2}\textbf{298.74} & \cellcolor{intercolor_2}\textbf{285.06} & \cellcolor{intercolor_2}\textbf{303.49} & \cellcolor{intercolor_2}\textbf{243.55} & \cellcolor{intercolor_2}\textbf{286.20} & \cellcolor{intercolor_2}\textbf{295.45} & \cellcolor{intercolor_2}\textbf{300.29} & \cellcolor{intercolor_2}\textbf{303.39} & \cellcolor{intercolor_2}\textbf{320.87}    \\ \hline \hline
\multirow{2}{*}{\rotatebox{0}{\footnotesize{$\Delta$}}}    & {$I_n$} & \textcolor{applegreen}{+19.96} & \textcolor{applegreen}{+13.12}  & \textcolor{applegreen}{+7.36} & \textcolor{applegreen}{+18.2} & \textcolor{downred}{-5.93} & \textcolor{applegreen}{+5.84} & \textcolor{applegreen}{+15.58} & \textcolor{applegreen}{+2.06} & \textcolor{applegreen}{+56.96} & \textcolor{applegreen}{+71.43} \\ 
                         
                         & {$I_l$} & \textcolor{applegreen}{+19.62} & \textcolor{applegreen}{+14.27} & \textcolor{applegreen}{+7.59} & \textcolor{applegreen}{+4.45} & \textcolor{applegreen}{+20.56} & \textcolor{applegreen}{+19.29} & \textcolor{applegreen}{+11.45} & \textcolor{applegreen}{+11.03} & \textcolor{applegreen}{+13.44} & \textcolor{applegreen}{+33.39} \\ \hline
\end{tabular}
\end{adjustbox}
\label{tab:EQQA}
\end{table}

A key intuitive finding reflected by Fig.~\ref{fig:PopQA} and Fig.~\ref{fig:EQ} is the superior performance of our method in long-context scenarios, particularly evident when $\mathcal{K}$ is high. As $\mathcal{K}$ increases, especially within the longer context setting ($I_l$), the $Acc.$ of our method consistently outperforms that of various backbone LLMs coupled with other context compression methods. This trend suggests that the concepts distilled by our method are supportive of reducing interference and enabling the LLMs to concentrate on key knowledge. Moreover, the positive values of $\Delta$ in Table~\ref{tab:PopQA} and Table~\ref{tab:EQQA} for the $I_l$ interval further underscore the improvement achieved by our framework over baseline methods when handling longer contexts. This observation emphasizes the effectiveness of the AMR-based concept distillation algorithm in capturing essential semantic information from supporting documents, thereby enabling LLMs to generate more accurate answers even when confronted with messy contexts.

When setting the bloom-560m model as the backbone LLMs, an interesting finding is that $\Delta$ exhibits negative trends in the $I_n$ interval of both datasets, while the SelCon does not perform ideally either. We hypothesize that this is due to the limitation of small-scale models to associate semantic scenarios through discrete concepts, which results in the model's inability to understand the core information expressed in the compressed supporting documents. Conversely, when coupling advanced LLMs, such as LLaMA-2, the contexts compressed by the proposed method and SelCon exhibit the most significant and second most significant enhancements to the LLMs, respectively. This observation likely arises from these large-scale models' superior contextual understanding capabilities, which corroborates our hypothesis. Regarding the improvements of $\Delta$ on $I_l$ interval of two datasets, our method's enhancement on the PopQA dataset is more pronounced. This is because PopQA was released recently, and its knowledge is less likely to be memorized by earlier models such as GPT-Neo and OPT. Moreover, the screening of long-tail knowledge further accentuates the unique scenario provided by PopQA, making it an ideal testbed for evaluating context compression methods.

The proposed AMR-based concept distillation method demonstrates clear advantages over generative compression methods of keyword extraction and summarization. While these methods utilise the LLMs to generate compressed representations and show competitive results in certain cases, they may inadvertently introduce noise or lose essential details during the compression process. Moreover, the generative nature of these methods makes them inherently difficult to control, even when provided with instructions as constraints. Consequently, the generated keywords and summaries may exhibit randomness, potentially deviating from the core concepts conveyed in the original supporting documents. In contrast, our framework leverages the inherent structured semantic representation of AMR to capture the core concepts explicitly. This semantic-level abstraction enables the framework to faithfully format the concepts to provide more reliable and informative support for the RAG process. 

Compared to the linguistics context compression baseline, SelCon, which identifies and prunes redundant content based on self-information computed at the lexical level, the proposed method based on the semantic level achieves superior results. SelCon's effectiveness depends on determining the right granularity for redundancy removal, making it sensitive to lexical unit choice. In contrast, our method takes a macro view by focusing on the semantic consistency carried by the AMR structure, making it insensitive to the delicate lexical bias. This characteristic enables it to be a reliable plug-and-play component in various RAG systems dealing with supporting documents containing irrelevant information and potential lexical errors. The robustness of the proposed framework is demonstrated by its consistent performance improvements across various LLMs. The experimental results on both datasets showcase the generalizability of our method, irrespective of the underlying LLM architecture. This finding suggests that the concept-based RAG framework can be effectively coupled with diverse LLMs, making it a versatile solution for enhancing inference performance in long-context scenarios.

\section{Conclusion and Future Research}

This paper introduces a novel concept-based RAG framework that utilizes AMR to distil essential concepts from long-context supporting documents, enabling LLMs to focus on the most supportive knowledge for accurate question-answering efficiently. The proposed AMR-based concept distillation algorithm systematically traverses the AMR graph to format key concept nodes with informative semantic features, transforming redundant supporting documents into a concise concept set. The proposed framework significantly enhances RAG performance compared with baselines comprising various backbone LLMs and context compression methods. To the best of our knowledge, this is the first work to augment RAG with AMR, offering a novel direction for integrating reliable structured semantic representations with RAG to handle tasks requiring high fidelity to the knowledge.

It has been demonstrated that the LLMs with fewer parameters within the proposed framework can also exhibit comparable or superior performance to larger models in certain cases. Consequently, it is plausible to speculate on the feasibility of employing small-scale LLMs solely equipped with the general natural language understanding capabilities, coupled with comprehensive and informative concept sets, to implement the lightweight Q\&A systems. This approach would alleviate the constraints imposed by the computational complexity of large-scale LLMs during their practical application and deployment. Exploring this possibility will be one of the focus of our future research.

\newpage

\bibliographystyle{plainnat}
\bibliography{reference}

\newpage

\appendix

\setcounter{figure}{0}
\setcounter{table}{0}
\renewcommand{\thefigure}{A\arabic{figure}}
\renewcommand{\thetable}{A\arabic{table}}

\section{AMR Examples}
\label{sec:AMR_example}

\begin{figure*}[htbp]
 \centering
 \includegraphics[width=1\textwidth]{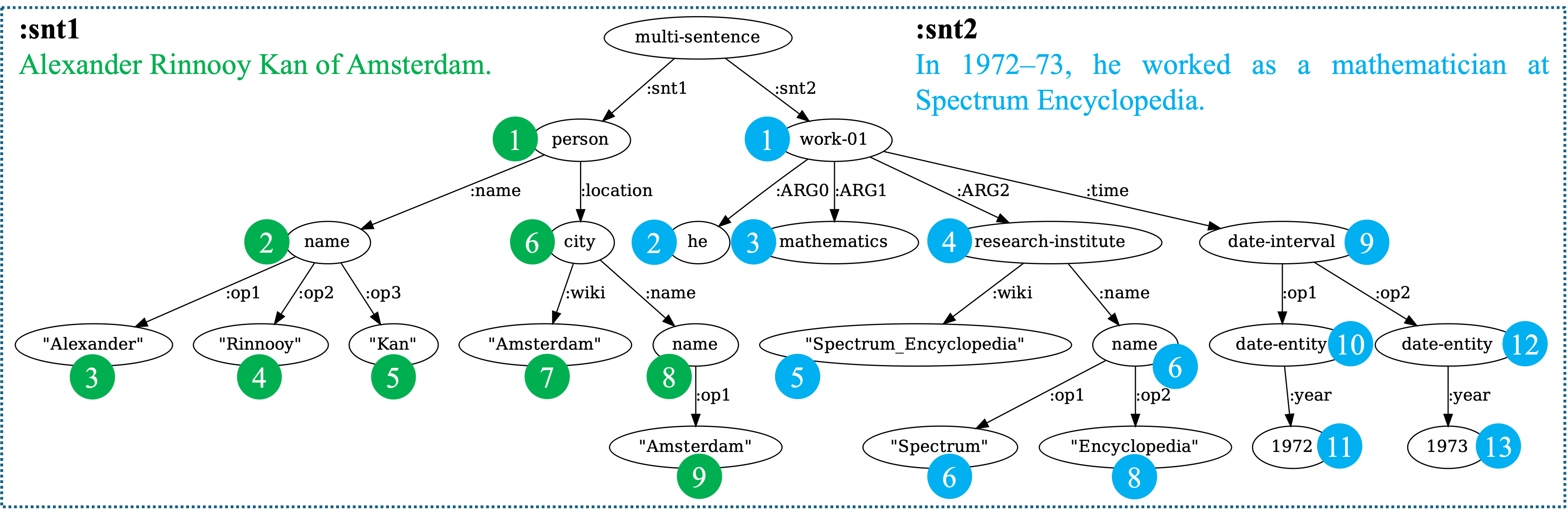}
 \caption{The node order of the DFS traversal within the AMR graph, with nodes marked in different colours for the two sentences. The marked nodes here are not the final distilled concepts.}
 \label{fig:AMR_DFS}
\end{figure*}

This appendix presents an example of an AMR graph parsed from the supporting document, "\texttt{Alexander Rinnooy Kan of Amsterdam. In 1972–73, he worked as a mathematician at Spectrum Encyclopedia}". Fig.~\ref{fig:AMR_DFS} showcases the DFS order for traversing the nodes in the AMR graph. Table~\ref{tab:AMR_tc} delineates the parsed AMR graph marked with the distilled concepts, showing that DFS traversal maintains the relative position of the adjacent concepts contained in the raw retrieved supporting document.

\begin{table}[htbp]
    \centering
    \caption{Example of the parsed AMR graph. The nodes carrying concepts are \textbf{bolded}, including "Alexander Rinnooy Kan, Amsterdam. work, mathematics, Spectrum Encyclopedia, 1972, 1973". The special roles, \colorbox{rolecolor_3}{\texttt{:wiki}}, \colorbox{rolecolor_2}{\texttt{:year}}, and \colorbox{rolecolor_1}{\texttt{:name}} are highlighted.}
    \label{tab:AMR_tc}
    \begin{tabular}{p{0.7\linewidth}}
    \hline
    \textbf{Supporting Document:} Alexander Rinnooy Kan of Amsterdam. In 1972–73, he worked as a mathematician at Spectrum Encyclopedia.  \\
        \hline
        \hspace{1cm}(m / multi-sentence \\
        \hspace{1.5cm}:snt1 (p / person \\
        \cellcolor{rolecolor_1}\hspace{2.5cm}:name (n / name \\
        \cellcolor{rolecolor_1}\hspace{3.5cm}:op1 \textbf{"Alexander"} \\
        \cellcolor{rolecolor_1}\hspace{3.5cm}:op2 \textbf{"Rinnooy"} \\
        \cellcolor{rolecolor_1}\hspace{3.5cm}:op3 \textbf{"Kan"}) \\
        \hspace{2.5cm}:location (c / city \\
        \cellcolor{rolecolor_3}\hspace{3.5cm}:wiki \textbf{"Amsterdam"} \\
        \cellcolor{rolecolor_1}\hspace{3.5cm}:name (n2 / name \\
        \cellcolor{rolecolor_1}\hspace{5cm}:op1 \textbf{"Amsterdam"}))) \\
        \hspace{1.5cm}:snt2 (w / \textbf{work}-01 \\
        \hspace{2.5cm}:ARG0 (h / he) \\
        \hspace{2.5cm}:ARG1 (m2 / \textbf{mathematics}) \\
        \hspace{2.5cm}:ARG2 (r / research-institute \\
        \cellcolor{rolecolor_3}\hspace{3.5cm}:wiki \textbf{"Spectrum\_Encyclopedia"} \\
        \cellcolor{rolecolor_1}\hspace{3.5cm}:name (n3 / name \\
        \cellcolor{rolecolor_1}\hspace{5cm}:op1 \textbf{"Spectrum"} \\
        \cellcolor{rolecolor_1}\hspace{5cm}:op2 \textbf{"Encyclopedia"})) \\
        \hspace{2.5cm}:time (d / date-interval \\
        \cellcolor{rolecolor_2}\hspace{3.5cm}:op1 (d2 / date-entity \\
        \cellcolor{rolecolor_2}\hspace{5cm}:year \textbf{1972}) \\
        \cellcolor{rolecolor_2}\hspace{3.5cm}:op2 (d3 / date-entity \\
        \cellcolor{rolecolor_2}\hspace{5cm}:year \textbf{1973})))) \\
        \hline
    \end{tabular}
\end{table}

\section{Prompts for Baselines}
\label{sec:prompt}

We refer to \citet{alpaca} to set up prompt templates for baselines with different instructions, as illustrated in Table~\ref{tab:prompt}.

\begin{table}[htbp]
\caption{The prompt for extracting and generating the keywords and summaries of supporting documents as baselines.}
\label{tab:prompt}
\centering
\begin{tabular}{p{13cm}}
\hline
Instruction (Keywords): \texttt{Extract a few keywords from the following content.}\\ 
Instruction (Summary): \texttt{Generate a short summary of the following content.}\\ \hline \hline
\begin{tabular}[c]{@{}p{13cm}@{}}Prompt = """Below is an instruction that describes a task, paired with an input that provides content.\\     \#\#\# Instruction: \{""" + \texttt{Instruction} + """\}\\    \#\#\# Input: \{""" + $\mathcal{D}$ + """\}\\     \#\#\# Response: """\end{tabular} \\ \hline
\end{tabular}
\end{table}

\section{Accuracy Details}
\label{sec:acc_details}

The detailed accuracy ($Acc.$) for the different concept compression methods on PopQA and EntityQuestions datasets are presented in Table~\ref{tab:aca_PopQA} and Table~\ref{tab:aca_EQQA}, respectively.

\begin{table}[htbp]
\centering
\caption{Accuracy ($Acc. \uparrow$) comparison on the PopQA dataset. The best results for each LLM with setting $\mathcal{K}$ are in \textbf{bold}, and the second best results are in \uline{underlined}. $\Delta$ here represents the difference between the methods of "Ours" and "Vanilla", and the \textcolor{applegreen}{increased} and \textcolor{downred}{decreased} $\Delta$ are marked differently. The best results for each of $\mathcal{K}$ are \textcolor{bestcolor}{marked}.}
\label{tab:aca_PopQA}
\begin{adjustbox}{width=1\linewidth}
\begin{tabular}{c|c|*{10}{c}}
\hline
LLMs & \diagbox{$\mathcal{C}$}{$\mathcal{K}$} & 1 & 2 & 3 & 4 & 5 & 6 & 7 & 8 & 9 & 10 \\
\hline
\multirow{6}{*}{\rotatebox{90}{GPT-Neo-1.3B}} & Vanilla & \textbf{47.69} & \textbf{71.38} & \textbf{69.19} & \textbf{67.56} & \uline{60.87} & \uline{73.51} & \uline{69.44} & \uline{72.15} & \uline{72.73} & \uline{80.00} \\
 & Keywords & 20.46 & 27.85 & 39.81 & 48.47 & 59.63 & 51.66 & 58.33 & 68.35 & 68.18 & 72.86 \\
 & Summary & 22.76 & 41.70 & 50.24 & 59.54 & 57.14 & 68.87 & 56.48 & 59.49 & \uline{72.73} & \uline{80.00} \\
 & SelCon & 21.14 & 34.74 & 39.57 & 41.60 & 52.17 & 56.29 & 47.22 & 58.23 & 68.18 & 71.43 \\
 & \cellcolor{markcolor}{Ours} & \cellcolor{markcolor}\uline{34.96} & \cellcolor{markcolor}\uline{55.62} & \cellcolor{markcolor}\uline{61.61} & \cellcolor{markcolor}\uline{64.89} & \cellcolor{markcolor}\textbf{64.60} & \cellcolor{markcolor}\textbf{77.48} & \cellcolor{markcolor}\textbf{75.93} & \cellcolor{markcolor}\textbf{77.22} & \cellcolor{markcolor}\textbf{83.33} & \cellcolor{markcolor}\textbf{94.29} \\
 & $\Delta$ & \textcolor{downred}{-12.73} & \textcolor{downred}{-15.76} & \textcolor{downred}{-7.58} & \textcolor{downred}{-2.67} & \textcolor{applegreen}{+3.73} & \textcolor{applegreen}{+3.97} & \textcolor{applegreen}{+6.49} & \textcolor{applegreen}{+5.07} & \textcolor{applegreen}{+10.60} & \textcolor{applegreen}{+14.29} \\
\hline
\multirow{6}{*}{\rotatebox{90}{GPT-Neo-2.7B}} & Vanilla & \textbf{58.13} & \textbf{76.13} & \textbf{71.56} & \textbf{73.66} & \uline{69.56} & \uline{72.19} & \uline{73.15} & 59.49 & \uline{65.15} & \textbf{82.86} \\
 & Keywords & 23.98 & 33.51 & 46.92 & 56.49 & 58.39 & 66.89 & 59.26 & 59.49 & 60.61 & 62.86 \\
 & Summary & 23.71 & 40.63 & 49.29 & 59.16 & 59.01 & 65.56 & 68.52 & 55.70 & 63.64 & 80.00 \\
 & SelCon & 33.88 & 48.58 & 46.45 & 57.25 & 62.73 & 59.60 & 58.33 & \uline{74.68} & 62.12 & 75.71 \\
 & \cellcolor{markcolor}{Ours} & \cellcolor{markcolor}\uline{43.09} & \cellcolor{markcolor}\uline{61.44} & \cellcolor{markcolor}\uline{64.45} & \cellcolor{markcolor}\uline{70.61} & \cellcolor{markcolor}\textbf{74.53} & \cellcolor{markcolor}\textbf{76.82} & \cellcolor{markcolor}\textbf{77.78} & \cellcolor{markcolor}\textbf{89.87} & \cellcolor{markcolor}\textbf{74.24} & \cellcolor{markcolor}\textbf{82.86} \\
 & $\Delta$ & \textcolor{downred}{-15.04} & \textcolor{downred}{-14.69} & \textcolor{downred}{-7.11} & \textcolor{downred}{-3.05} & \textcolor{applegreen}{+4.97} & \textcolor{applegreen}{+4.63} & \textcolor{applegreen}{+4.63} & \textcolor{applegreen}{+30.38} & \textcolor{applegreen}{+9.09} & 0.00 \\
\hline
\multirow{6}{*}{\rotatebox{90}{OPT-1.3b}} & Vanilla & \textbf{62.47} & \textbf{75.52} & \textbf{63.51} & \uline{76.34} & 70.81 & 76.82 & 68.52 & 74.68 & 74.24 & 90.00 \\
 & Keywords & 25.07 & 33.89 & 49.05 & 58.78 & 68.94 & 67.55 & \uline{75.00} & 74.68 & 74.24 & 80.00 \\
 & Summary & 27.91 & 47.05 & 59.48 & 69.08 & \textbf{72.67} & \textbf{81.46} & 73.15 & 74.68 & \textbf{81.82} & \uline{92.86} \\
 & SelCon & 40.11 & 56.77 & 58.06 & 67.18 & 65.22 & 70.20 & 72.22 & \uline{82.28} & \uline{78.79} & 90.00 \\
 & \cellcolor{markcolor}{Ours} & \cellcolor{markcolor}\uline{47.56} & \cellcolor{markcolor}\uline{63.12} & \cellcolor{markcolor}\textbf{63.51} & \cellcolor{markcolor}\textbf{76.72} & \cellcolor{markcolor}\textbf{72.67} & \cellcolor{markcolor}\uline{78.81} & \cellcolor{markcolor}\textbf{77.78} & \cellcolor{markcolor}\textbf{87.34} & \cellcolor{markcolor}77.27 & \cellcolor{markcolor}\textbf{95.71} \\
 & $\Delta$ & \textcolor{downred}{-14.91} & \textcolor{downred}{-12.40} & 0.00 & \textcolor{applegreen}{+0.38} & \textcolor{applegreen}{+1.86} & \textcolor{applegreen}{+1.99} & \textcolor{applegreen}{+9.26} & \textcolor{applegreen}{+12.66} & \textcolor{applegreen}{+3.03} & \textcolor{applegreen}{+5.71} \\
\hline
\multirow{6}{*}{\rotatebox{90}{OPT-2.7b}} & Vanilla & \textbf{62.33} & \textbf{77.43} & \textbf{72.99} & \textbf{77.48} & 66.45 & \uline{78.80} & \uline{77.78} & 79.75 & \uline{80.30} & \uline{90.00} \\
 & Keywords & 27.51 & 32.44 & 49.05 & 62.60 & \uline{70.81} & 74.17 & 68.52 & 78.48 & 77.27 & 88.57 \\
 & Summary & 26.29 & 46.14 & 58.06 & 66.41 & 63.98 & 74.17 & 71.30 & 65.82 & 72.72 & 87.14 \\
 & SelCon & 33.74 & 51.34 & 50.24 & 60.69 & 65.22 & 78.15 & 75.93 & \uline{84.81} & \uline{80.30} & 87.14 \\
 & \cellcolor{markcolor}{Ours} & \cellcolor{markcolor}\uline{49.86} & \cellcolor{markcolor}\uline{65.11} & \cellcolor{markcolor}\uline{68.96} & \cellcolor{markcolor}\uline{71.76} & \cellcolor{markcolor}\textbf{72.67} & \cellcolor{markcolor}\textbf{81.46} & \cellcolor{markcolor}\textbf{83.33} & \cellcolor{markcolor}\textbf{89.87} & \cellcolor{markcolor}\textbf{81.81} & \cellcolor{markcolor}\textcolor{bestcolor}{\textbf{97.14}} \\
 & $\Delta$ & \textcolor{downred}{-12.47} & \textcolor{downred}{-12.32} & \textcolor{downred}{-4.03} & \textcolor{downred}{-5.72} & \textcolor{applegreen}{+6.22} & \textcolor{applegreen}{+2.66} & \textcolor{applegreen}{+5.55} & \textcolor{applegreen}{+10.12} & \textcolor{applegreen}{+1.51} & \textcolor{applegreen}{+7.14} \\
\hline
\multirow{6}{*}{\rotatebox{90}{Bloom-560m}} & Vanilla & \textbf{52.30} & \textbf{64.58} & \textbf{63.03} & \textbf{68.32} & \uline{67.08} & \uline{71.52} & \uline{73.14} & \uline{77.22} & 66.67 & \uline{84.29} \\
 & Keywords & 21.68 & 29.23 & 46.21 & 51.15 & 57.76 & 64.24 & 63.89 & 73.42 & 66.67 & 78.57 \\
 & Summary & 24.39 & 42.16 & 53.79 & 63.36 & 61.49 & \textbf{72.82} & \textbf{75.00} & 75.95 & \uline{75.76} & 82.86 \\
 & SelCon & 22.90 & 30.68 & 34.83 & 40.46 & 48.45 & 54.30 & 50.93 & 63.29 & 54.55 & 68.57 \\
 & \cellcolor{markcolor}{Ours} & \cellcolor{markcolor}\uline{35.91} & \cellcolor{markcolor}\uline{49.12} & \cellcolor{markcolor}\uline{54.50} & \cellcolor{markcolor}\textbf{68.32} & \cellcolor{markcolor}\textbf{68.32} & \cellcolor{markcolor}\uline{71.52} & \cellcolor{markcolor}67.59 & \cellcolor{markcolor}\textbf{83.54} & \cellcolor{markcolor}\textbf{80.30} & \cellcolor{markcolor}\textbf{94.29} \\
 & $\Delta$ & \textcolor{downred}{-16.39} & \textcolor{downred}{-15.46} & \textcolor{downred}{-8.53} & 0.00 & \textcolor{applegreen}{+1.24} & 0.00 & \textcolor{downred}{-5.55} & \textcolor{applegreen}{+6.32} & \textcolor{applegreen}{+13.63} & \textcolor{applegreen}{+10.00} \\
\hline
\multirow{6}{*}{\rotatebox{90}{Bloom-1b1}} & Vanilla & \textbf{63.01} & \textbf{75.06} & \textbf{70.62} & \textbf{77.10} & 73.29 & \uline{79.47} & 75.93 & \uline{87.34} & 78.79 & 87.14 \\
 & Keywords & 26.15 & 39.40 & 56.87 & 64.89 & 72.67 & 72.19 & 75.93 & \uline{87.34} & \uline{83.33} & \uline{90.00} \\
 & Summary & 27.10 & 46.37 & 56.87 & 70.61 & 69.57 & 76.16 & \uline{78.70} & 73.42 & 80.30 & 85.71 \\
 & SelCon & \uline{45.53} & \uline{59.68} & 54.74 & 66.41 & \uline{75.16} & 74.83 & \uline{78.70} & 77.22 & 80.30 & \uline{90.00} \\
 & \cellcolor{markcolor}{Ours} & \cellcolor{markcolor}42.55 & \cellcolor{markcolor}58.68 & \cellcolor{markcolor}\uline{65.64} & \cellcolor{markcolor}\uline{71.76} & \cellcolor{markcolor}\textcolor{bestcolor}{\textbf{79.50}} & \cellcolor{markcolor}\textbf{83.44} & \cellcolor{markcolor}\textbf{83.33} & \cellcolor{markcolor}\textbf{89.87} & \cellcolor{markcolor}\textbf{86.36} & \cellcolor{markcolor}\textbf{92.86} \\
 & $\Delta$ & \textcolor{downred}{-20.46} & \textcolor{downred}{-16.38} & \textcolor{downred}{-4.98} & \textcolor{downred}{-5.34} & \textcolor{applegreen}{+6.21} & \textcolor{applegreen}{+3.97} & \textcolor{applegreen}{+7.40} & \textcolor{applegreen}{+2.53} & \textcolor{applegreen}{+7.57} & \textcolor{applegreen}{+5.72} \\
\hline
\multirow{6}{*}{\rotatebox{90}{Bloom-1b7}} & Vanilla & \textbf{65.04} & \textbf{78.35} & \textbf{76.30} & \textcolor{bestcolor}{\textbf{79.77}} & \uline{73.29} & \uline{81.46} & \uline{78.70} & 82.28 & 80.30 & 88.57 \\
 & Keywords & 28.05 & 39.86 & 55.21 & 65.65 & 72.67 & 72.85 & \uline{78.70} & \uline{92.41} & \uline{83.33} & \uline{94.29} \\
 & Summary & 29.54 & 46.98 & 59.24 & 71.76 & 70.19 & 76.82 & \uline{78.70} & 88.61 & \uline{83.33} & \uline{94.29} \\
 & SelCon & 40.51 & 57.15 & 54.74 & 67.18 & 63.35 & 73.51 & 71.30 & 82.28 & 74.24 & 84.29 \\
 & \cellcolor{markcolor}{Ours} & \cellcolor{markcolor}\uline{47.43} & \cellcolor{markcolor}\uline{61.74} & \cellcolor{markcolor}\uline{66.82} & \cellcolor{markcolor}\uline{75.19} & \cellcolor{markcolor}\textbf{74.53} & \cellcolor{markcolor}\textbf{82.78} & \cellcolor{markcolor}\textbf{84.26} & \cellcolor{markcolor}\textcolor{bestcolor}{\textbf{94.94}} & \cellcolor{markcolor}\textbf{86.36} & \cellcolor{markcolor}\textcolor{bestcolor}{\textbf{97.14}} \\
 & $\Delta$ & \textcolor{downred}{-17.61} & \textcolor{downred}{-16.61} & \textcolor{downred}{-9.48} & \textcolor{downred}{-4.58} & \textcolor{applegreen}{+1.24} & \textcolor{applegreen}{+1.32} & \textcolor{applegreen}{+5.56} & \textcolor{applegreen}{+12.66} & \textcolor{applegreen}{+6.06} & \textcolor{applegreen}{+8.57} \\
\hline
\multirow{6}{*}{\rotatebox{90}{Bloom-3b}} & Vanilla & \textbf{64.36} & \textcolor{bestcolor}{\textbf{78.81}} & \textbf{73.70} & \textbf{74.05} & \uline{70.80} & 72.85 & \uline{79.62} & 78.48 & 74.24 & 74.29 \\
 & Keywords & 28.86 & 40.24 & 56.87 & 64.12 & 70.19 & 72.85 & 76.85 & 83.54 & 77.27 & 88.57 \\
 & Summary & 26.29 & 46.60 & 56.40 & 70.22 & 69.57 & \uline{75.50} & 72.22 & 64.56 & 75.76 & \uline{94.29} \\
 & SelCon & 46.34 & 58.45 & 59.48 & 66.41 & 68.32 & 71.52 & 68.52 & \textbf{87.34} & \uline{80.30} & 84.29 \\
 & \cellcolor{markcolor}{Ours} & \cellcolor{markcolor}\uline{50.00} & \cellcolor{markcolor}\uline{60.67} & \cellcolor{markcolor}\uline{64.69} & \cellcolor{markcolor}\uline{72.52} & \cellcolor{markcolor}\textbf{74.53} & \cellcolor{markcolor}\textbf{79.47} & \cellcolor{markcolor}\textbf{86.11} & \cellcolor{markcolor}\uline{84.81} & \cellcolor{markcolor}\textbf{84.85} & \cellcolor{markcolor}\textcolor{bestcolor}{\textbf{97.14}} \\
 & $\Delta$ & \textcolor{downred}{-14.36} & \textcolor{downred}{-18.14} & \textcolor{downred}{-9.01} & \textcolor{downred}{-1.53} & \textcolor{applegreen}{+3.73} & \textcolor{applegreen}{+6.62} & \textcolor{applegreen}{+6.49} & \textcolor{applegreen}{+6.33} & \textcolor{applegreen}{+10.61} & \textcolor{applegreen}{+22.85} \\
\hline
\multirow{6}{*}{\rotatebox{90}{LLaMA-2-c-7}} & Vanilla & 57.32 & 69.93 & 62.09 & 72.90 & 70.81 & 81.45 & \uline{85.19} & 78.48 & 86.36 & \uline{92.86} \\
 & Keywords & 35.91 & 49.50 & 64.93 & 67.56 & 63.35 & 78.15 & 82.41 & 77.22 & 84.85 & 85.71 \\
 & Summary & 29.54 & 42.77 & 51.66 & 67.18 & 60.87 & 71.52 & 77.78 & 68.35 & 68.18 & 82.86 \\
 & SelCon & \uline{58.81} & \textbf{74.37} & \uline{73.93} & \textbf{79.00} & \uline{72.05} & \uline{86.09} & 82.41 & \uline{83.54} & \textcolor{bestcolor}{\textbf{89.39}} & 91.43 \\
 & \cellcolor{markcolor}{Ours} & \cellcolor{markcolor}\textcolor{bestcolor}{\textbf{67.74}} & \cellcolor{markcolor}\uline{74.06} & \cellcolor{markcolor}\textbf{74.64} & \cellcolor{markcolor}\uline{76.72} & \cellcolor{markcolor}\textbf{78.26} & \cellcolor{markcolor}\textcolor{bestcolor}{\textbf{87.42}} & \cellcolor{markcolor}\textcolor{bestcolor}{\textbf{89.81}} & \cellcolor{markcolor}\textbf{86.08} & \cellcolor{markcolor}\textcolor{bestcolor}{\textbf{89.39}} & \cellcolor{markcolor}\textcolor{bestcolor}{\textbf{97.14}} \\
 & $\Delta$ & \textcolor{applegreen}{+10.42} & \textcolor{applegreen}{+4.13} & \textcolor{applegreen}{+12.55} & \textcolor{applegreen}{+3.82} & \textcolor{applegreen}{+7.45} & \textcolor{applegreen}{+5.97} & \textcolor{applegreen}{+4.62} & \textcolor{applegreen}{+7.60} & \textcolor{applegreen}{+3.03} & \textcolor{applegreen}{+4.28} \\
\hline
\multirow{6}{*}{\rotatebox{90}{LLaMA-2-c-13}} & Vanilla & 59.21 & \textbf{77.74} & 68.96 & 74.43 & 68.94 & 79.47 & 75.93 & 74.68 & 77.27 & 90.00 \\
 & Keywords & 37.40 & 50.11 & 59.48 & 66.03 & 67.08 & 79.47 & 75.00 & 72.15 & \textbf{83.33} & \uline{91.43} \\
 & Summary & 28.86 & 46.29 & 52.13 & 59.16 & 58.39 & 62.91 & 72.22 & 64.56 & 80.30 & 85.71 \\
 & SelCon & \textbf{64.36} & \uline{74.90} & \uline{74.64} & \uline{75.57} & \uline{71.43} & \textbf{84.11} & \uline{82.41} & \uline{83.54} & 78.79 & \uline{91.43} \\
 & \cellcolor{markcolor}{Ours} & \cellcolor{markcolor}\uline{62.06} & \cellcolor{markcolor}73.14 & \cellcolor{markcolor}\textcolor{bestcolor}{\textbf{76.54}} & \cellcolor{markcolor}\textbf{78.63} & \cellcolor{markcolor}\textbf{75.78} & \cellcolor{markcolor}\textbf{84.11} & \cellcolor{markcolor}\textbf{84.26} & \cellcolor{markcolor}\textbf{84.81} & \cellcolor{markcolor}\uline{81.82} & \cellcolor{markcolor}\textbf{92.86} \\
 & $\Delta$ & \textcolor{applegreen}{+2.85} & \textcolor{downred}{-4.60} & \textcolor{applegreen}{+7.58} & \textcolor{applegreen}{+4.20} & \textcolor{applegreen}{+6.84} & \textcolor{applegreen}{+4.64} & \textcolor{applegreen}{+8.33} & \textcolor{applegreen}{+10.13} & \textcolor{applegreen}{+4.55} & \textcolor{applegreen}{+2.86} \\
\hline

\end{tabular}
\end{adjustbox}
\end{table}

\begin{table}[htbp]
\centering
\caption{Accuracy ($Acc. \uparrow$) comparison the EntityQuestions dataset. The definitions of symbols are the same with Table~\ref{tab:aca_PopQA}.}
\label{tab:aca_EQQA}
\begin{adjustbox}{width=1\linewidth}
\begin{tabular}{c|c|*{10}{c}}
\hline
LLMs & \diagbox{$\mathcal{C}$}{$\mathcal{K}$} & 1 & 2 & 3 & 4 & 5 & 6 & 7 & 8 & 9 & 10 \\
\hline
\multirow{6}{*}{\rotatebox{90}{GPT-Neo-1.3B}} & Vanilla & \textbf{42.43} & \textbf{53.15} & \textbf{58.34} & \uline{57.27} & \uline{61.49} & \uline{65.15} & \uline{59.18} & \uline{57.83} & \uline{64.42} & \uline{66.99} \\
 & Keywords & 19.39 & 26.62 & 30.15 & 30.40 & 30.75 & 36.36 & 31.12 & 30.12 & 40.49 & 30.10 \\
 & Summary & 26.21 & 34.07 & 38.51 & 36.78 & 41.19 & 46.21 & 43.88 & 37.95 & 47.85 & 54.36 \\
 & SelCon & 21.13 & 27.68 & 28.81 & 31.28 & 34.63 & 41.67 & 32.65 & 32.53 & 39.26 & 38.83 \\
 & \cellcolor{markcolor}{Ours} & \cellcolor{markcolor}\uline{36.74} & \cellcolor{markcolor}\uline{51.38} & \cellcolor{markcolor}\uline{57.61} & \cellcolor{markcolor}\textbf{59.91} & \cellcolor{markcolor}\textbf{63.58} & \cellcolor{markcolor}\textbf{67.05} & \cellcolor{markcolor}\textbf{65.31} & \cellcolor{markcolor}\textbf{62.65} & \cellcolor{markcolor}\textbf{71.17} & \cellcolor{markcolor}\textbf{68.93} \\
 & $\Delta$ & \textcolor{downred}{-5.69} & \textcolor{downred}{-1.77} & \textcolor{downred}{-0.73} & \textcolor{applegreen}{+2.64} & \textcolor{applegreen}{+2.09} & \textcolor{applegreen}{+1.90} & \textcolor{applegreen}{+6.13} & \textcolor{applegreen}{+4.82} & \textcolor{applegreen}{+6.75} & \textcolor{applegreen}{+1.94} \\
\hline
\multirow{6}{*}{\rotatebox{90}{GPT-Neo-2.7B}} & Vanilla & \textbf{51.05} & \textbf{61.58} & \textbf{65.22} & \textbf{67.18} & \uline{65.67} & \uline{70.83} & \uline{69.90} & \uline{70.48} & \uline{71.78} & \uline{73.79} \\
 & Keywords & 26.70 & 32.74 & 40.00 & 42.51 & 38.51 & 51.51 & 42.86 & 34.34 & 42.33 & 43.69 \\
 & Summary & 25.91 & 36.73 & 40.60 & 42.29 & 45.07 & 48.11 & 48.47 & 53.01 & 55.21 & 48.54 \\
 & SelCon & 30.04 & 38.86 & 40.75 & 44.27 & 46.27 & 50.00 & 44.90 & 43.98 & 53.99 & 54.36 \\
 & \cellcolor{markcolor}{Ours} & \cellcolor{markcolor}\uline{48.95} & \cellcolor{markcolor}\uline{58.03} & \cellcolor{markcolor}\uline{64.33} & \cellcolor{markcolor}\uline{67.00} & \cellcolor{markcolor}\textbf{69.25} & \cellcolor{markcolor}\textbf{72.72} & \cellcolor{markcolor}\textbf{72.96} & \cellcolor{markcolor}\textbf{71.08} & \cellcolor{markcolor}\textbf{78.53} & \cellcolor{markcolor}\textcolor{bestcolor}{\textbf{79.61}} \\
 & $\Delta$ & \textcolor{downred}{-2.10} & \textcolor{downred}{-3.55} & \textcolor{downred}{-0.89} & \textcolor{downred}{-0.18} & \textcolor{applegreen}{+3.58} & \textcolor{applegreen}{+1.89} & \textcolor{applegreen}{+3.06} & \textcolor{applegreen}{+0.60} & \textcolor{applegreen}{+6.75} & \textcolor{applegreen}{+5.82} \\
\hline
\multirow{6}{*}{\rotatebox{90}{OPT-1.3b}} & Vanilla & \textbf{54.04} & \textbf{65.66} & \textbf{65.67} & \uline{65.64} & \uline{65.07} & \uline{71.97} & \uline{66.33} & \textbf{69.88} & \uline{71.78} & \uline{66.99} \\
 & Keywords & 29.68 & 33.54 & 40.60 & 43.61 & 47.46 & 54.55 & 51.53 & 50.60 & 44.79 & 43.69 \\
 & Summary & 31.18 & 42.86 & 49.10 & 51.98 & 55.22 & 62.50 & 58.67 & 63.25 & 69.33 & 66.02 \\
 & SelCon & 36.27 & 44.54 & 45.37 & 49.12 & 54.03 & 56.82 & 55.10 & 52.41 & 63.80 & 64.08 \\
 & \cellcolor{markcolor}{Ours} & \cellcolor{markcolor}\uline{49.13} & \cellcolor{markcolor}\uline{61.93} & \cellcolor{markcolor}\uline{65.52} & \cellcolor{markcolor}\textbf{67.62} & \cellcolor{markcolor}\textbf{68.06} & \cellcolor{markcolor}\textbf{74.24} & \cellcolor{markcolor}\textbf{70.41} & \cellcolor{markcolor}\uline{65.66} & \cellcolor{markcolor}\textbf{75.46} & \cellcolor{markcolor}\textbf{72.82} \\
 & $\Delta$ & \textcolor{downred}{-4.91} & \textcolor{downred}{-3.73} & \textcolor{downred}{-0.15} & \textcolor{applegreen}{+1.98} & \textcolor{applegreen}{+2.99} & \textcolor{applegreen}{+2.27} & \textcolor{applegreen}{+4.08} & \textcolor{downred}{-4.22} & \textcolor{applegreen}{+3.68} & \textcolor{applegreen}{+5.83} \\
\hline
\multirow{6}{*}{\rotatebox{90}{OPT-2.7b}} & Vanilla & \textbf{57.87} & \textbf{67.70} & \textbf{70.44} & \uline{68.50} & \uline{69.85} & \uline{77.65} & \textbf{73.47} & \uline{73.49} & \uline{79.75} & \uline{66.99} \\
 & Keywords & 33.51 & 36.82 & 42.69 & 49.12 & 50.15 & 57.58 & 52.04 & 54.22 & 57.06 & 46.60 \\
 & Summary & 31.11 & 42.77 & 49.70 & 47.58 & 51.94 & 56.06 & 53.57 & 53.01 & 53.99 & 44.66 \\
 & SelCon & 35.43 & 44.19 & 46.87 & 49.78 & 54.63 & 62.88 & 52.04 & 52.41 & 59.51 & 56.31 \\
 & \cellcolor{markcolor}{Ours} & \cellcolor{markcolor}\uline{55.36} & \cellcolor{markcolor}\uline{66.99} & \cellcolor{markcolor}\uline{68.81} & \cellcolor{markcolor}\textbf{72.91} & \cellcolor{markcolor}\textbf{72.83} & \cellcolor{markcolor}\textbf{79.55} & \cellcolor{markcolor}\uline{67.86} & \cellcolor{markcolor}\textbf{75.90} & \cellcolor{markcolor}\textbf{81.60} & \cellcolor{markcolor}\textbf{76.70} \\
 & $\Delta$ & \textcolor{downred}{-2.51} & \textcolor{downred}{-0.71} & \textcolor{downred}{-1.63} & \textcolor{applegreen}{+4.41} & \textcolor{applegreen}{+2.98} & \textcolor{applegreen}{+1.90} & \textcolor{downred}{-5.61} & \textcolor{applegreen}{+2.41} & \textcolor{applegreen}{+1.85} & \textcolor{applegreen}{+9.71} \\
\hline
\multirow{6}{*}{\rotatebox{90}{Bloom-560m}} & Vanilla & \textbf{39.86} & \textbf{50.40} & \textbf{53.58} & \textbf{54.63} & \textbf{57.31} & \uline{60.23} & \uline{52.55} & \uline{51.81} & \uline{58.90} & \uline{59.22} \\
 & Keywords & 20.89 & 27.15 & 34.03 & 31.72 & 29.25 & 37.50 & 37.24 & 32.53 & 38.65 & 33.01 \\
 & Summary & 24.18 & 35.40 & 42.09 & 40.75 & 43.58 & 53.41 & 42.35 & 50.60 & 44.79 & 46.60 \\
 & SelCon & 19.87 & 23.96 & 20.90 & 15.42 & 22.09 & 29.55 & 26.53 & 24.10 & 29.45 & 27.18 \\
 & \cellcolor{markcolor}{Ours} & \cellcolor{markcolor}\uline{32.91} & \cellcolor{markcolor}\uline{44.90} & \cellcolor{markcolor}\uline{48.36} & \cellcolor{markcolor}\uline{47.14} & \cellcolor{markcolor}\uline{49.85} & \cellcolor{markcolor}\textbf{65.53} & \cellcolor{markcolor}\textbf{57.65} & \cellcolor{markcolor}\textbf{54.22} & \cellcolor{markcolor}\textbf{66.87} & \cellcolor{markcolor}\textbf{64.08} \\
 & $\Delta$ & \textcolor{downred}{-6.95} & \textcolor{downred}{-5.50} & \textcolor{downred}{-5.22} & \textcolor{downred}{-7.49} & \textcolor{downred}{-7.46} & \textcolor{applegreen}{+5.30} & \textcolor{applegreen}{+5.10} & \textcolor{applegreen}{+2.41} & \textcolor{applegreen}{+7.97} & \textcolor{applegreen}{+4.86} \\
\hline
\multirow{6}{*}{\rotatebox{90}{Bloom-1b1}} & Vanilla & \textbf{50.75} & \textbf{62.64} & \textbf{64.78} & \textbf{70.26} & \textbf{68.36} & \uline{73.11} & \uline{63.27} & \uline{63.65} & \uline{69.94} & \uline{66.99} \\
 & Keywords & 26.81 & 38.24 & 43.73 & 52.20 & 44.78 & 57.20 & 53.06 & 49.40 & 50.31 & 52.43 \\
 & Summary & 30.46 & 41.08 & 44.18 & 44.93 & 47.76 & 50.00 & 48.98 & 50.60 & 46.63 & 52.43 \\
 & SelCon & 35.73 & 45.70 & 46.87 & 52.64 & 52.24 & 60.23 & 55.10 & 50.00 & 53.99 & 51.46 \\
 & \cellcolor{markcolor}{Ours} & \cellcolor{markcolor}\uline{43.87} & \cellcolor{markcolor}\uline{58.74} & \cellcolor{markcolor}\uline{63.88} & \cellcolor{markcolor}\uline{66.96} & \cellcolor{markcolor}\uline{66.27} & \cellcolor{markcolor}\textbf{73.48} & \cellcolor{markcolor}\textbf{68.37} & \cellcolor{markcolor}\textbf{71.08} & \cellcolor{markcolor}\textbf{71.17} & \cellcolor{markcolor}\textbf{77.67} \\
 & $\Delta$ & \textcolor{downred}{-6.88} & \textcolor{downred}{-3.90} & \textcolor{downred}{-0.90} & \textcolor{downred}{-3.30} & \textcolor{downred}{-2.09} & \textcolor{applegreen}{+0.37} & \textcolor{applegreen}{+5.10} & \textcolor{applegreen}{+7.43} & \textcolor{applegreen}{+1.23} & \textcolor{applegreen}{+10.68} \\
\hline
\multirow{6}{*}{\rotatebox{90}{Bloom-1b7}} & Vanilla & \textbf{51.89} & \textbf{64.24} & \textbf{64.63} & \uline{67.18} & \uline{66.87} & \uline{71.97} & \uline{68.37} & \uline{70.48} & \uline{71.78} & \uline{74.76} \\
 & Keywords & 29.74 & 43.03 & 45.82 & 51.76 & 49.25 & 65.15 & 57.14 & 53.01 & 58.90 & 52.43 \\
 & Summary & 31.18 & 43.48 & 47.61 & 50.00 & 58.51 & 59.09 & 62.76 & 62.65 & 68.71 & 66.99 \\
 & SelCon & 34.41 & 43.66 & 44.93 & 50.22 & 50.15 & 55.68 & 54.08 & 47.59 & 58.28 & 59.22 \\
 & \cellcolor{markcolor}{Ours} & \cellcolor{markcolor}\uline{50.87} & \cellcolor{markcolor}\uline{62.29} & \cellcolor{markcolor}\uline{63.58} & \cellcolor{markcolor}\textbf{67.84} & \cellcolor{markcolor}\textbf{72.24} & \cellcolor{markcolor}\textbf{75.38} & \cellcolor{markcolor}\textbf{71.94} & \cellcolor{markcolor}\textbf{73.49} & \cellcolor{markcolor}\textbf{73.01} & \cellcolor{markcolor}\textbf{78.64} \\
 & $\Delta$ & \textcolor{downred}{-1.02} & \textcolor{downred}{-1.95} & \textcolor{downred}{-1.05} & \textcolor{applegreen}{+0.66} & \textcolor{applegreen}{+5.37} & \textcolor{applegreen}{+3.41} & \textcolor{applegreen}{+3.57} & \textcolor{applegreen}{+3.01} & \textcolor{applegreen}{+1.23} & \textcolor{applegreen}{+3.88} \\
\hline
\multirow{6}{*}{\rotatebox{90}{Bloom-3b}} & Vanilla & \textbf{51.94} & \textbf{62.73} & \textbf{67.61} & \textbf{68.06} & \uline{68.36} & \uline{74.62} & \uline{69.90} & \uline{72.89} & \uline{71.78} & \uline{74.76} \\
 & Keywords & 31.60 & 41.08 & 47.76 & 52.20 & 51.94 & 60.98 & 58.16 & 51.20 & 58.28 & 50.49 \\
 & Summary & 30.10 & 38.69 & 46.72 & 46.92 & 48.06 & 56.44 & 47.96 & 52.41 & 54.60 & 57.28 \\
 & SelCon & 42.97 & 49.16 & 54.18 & 57.49 & 61.49 & 67.80 & 60.20 & 60.24 & 70.55 & 73.79 \\
 & \cellcolor{markcolor}{Ours} & \cellcolor{markcolor}\uline{49.97} & \cellcolor{markcolor}\uline{58.65} & \cellcolor{markcolor}\uline{64.03} & \cellcolor{markcolor}\uline{65.86} & \cellcolor{markcolor}\textbf{69.85} & \cellcolor{markcolor}\textbf{75.38} & \cellcolor{markcolor}\textbf{71.94} & \cellcolor{markcolor}\textbf{77.11} & \cellcolor{markcolor}\textbf{74.23} & \cellcolor{markcolor}\textbf{78.64} \\
 & $\Delta$ & \textcolor{downred}{-1.97} & \textcolor{downred}{-4.08} & \textcolor{downred}{-3.58} & \textcolor{downred}{-2.20} & \textcolor{applegreen}{+1.49} & \textcolor{applegreen}{+0.76} & \textcolor{applegreen}{+2.04} & \textcolor{applegreen}{+4.22} & \textcolor{applegreen}{+2.45} & \textcolor{applegreen}{+3.88} \\
\hline
\multirow{6}{*}{\rotatebox{90}{LLaMA-2-c-7}} & Vanilla & 49.67 & 51.73 & 57.76 & 66.30 & \uline{70.15} & \uline{75.00} & \textbf{73.47} & \uline{74.39} & \uline{76.69} & 73.79 \\
 & Keywords & 41.53 & 56.52 & 55.67 & 62.33 & 65.37 & 72.35 & 70.41 & 70.48 & 74.85 & 71.84 \\
 & Summary & 36.39 & 40.73 & 42.54 & 44.71 & 44.18 & 50.00 & 49.49 & 50.00 & 60.12 & 51.46 \\
 & SelCon & \uline{50.99} & \uline{62.91} & \uline{65.22} & \uline{68.50} & 68.96 & \uline{75.00} & \uline{72.45} & 71.08 & 75.46 & \textbf{77.67} \\
 & \cellcolor{markcolor}{Ours} & \cellcolor{markcolor}\textbf{60.80} & \cellcolor{markcolor}\textbf{70.45} & \cellcolor{markcolor}\textbf{71.94} & \cellcolor{markcolor}\textbf{75.33} & \cellcolor{markcolor}\textbf{74.03} & \cellcolor{markcolor}\textbf{77.27} & \cellcolor{markcolor}\uline{72.45} & \cellcolor{markcolor}\textbf{75.30} & \cellcolor{markcolor}\textbf{79.14} & \cellcolor{markcolor}\uline{75.73} \\
 & $\Delta$ & \textcolor{applegreen}{+11.13} & \textcolor{applegreen}{+18.72} & \textcolor{applegreen}{+14.18} & \textcolor{applegreen}{+9.03} & \textcolor{applegreen}{+3.88} & \textcolor{applegreen}{+2.27} & \textcolor{downred}{-1.02} & \textcolor{applegreen}{+0.91} & \textcolor{applegreen}{+2.45} & \textcolor{applegreen}{+1.94} \\
\hline
\multirow{6}{*}{\rotatebox{90}{LLaMA-2-c-13}} & Vanilla & 52.36 & \uline{69.48} & \uline{71.04} & \uline{70.93} & 69.25 & \uline{75.76} & 67.35 & 69.28 & 76.07 & 73.79 \\
 & Keywords & 41.77 & 52.97 & 55.07 & 61.01 & 65.67 & 68.94 & 72.96 & 69.28 & 72.39 & 70.87 \\
 & Summary & 36.80 & 45.06 & 44.18 & 42.07 & 42.39 & 46.59 & 52.04 & 57.23 & 60.74 & 59.22 \\
 & SelCon & \uline{52.54} & 65.04 & 69.25 & 68.50 & \uline{70.45} & 73.86 & \uline{74.49} & \uline{75.30} & \uline{78.53} & \textbf{78.64} \\
 & \cellcolor{markcolor}{Ours} & \cellcolor{markcolor}\textcolor{bestcolor}{\textbf{63.61}} & \cellcolor{markcolor}\textcolor{bestcolor}{\textbf{74.37}} & \cellcolor{markcolor}\textcolor{bestcolor}{\textbf{74.18}} & \cellcolor{markcolor}\textcolor{bestcolor}{\textbf{81.72}} & \cellcolor{markcolor}\textcolor{bestcolor}{\textbf{80.00}} & \cellcolor{markcolor}\textcolor{bestcolor}{\textbf{81.44}} & \cellcolor{markcolor}\textcolor{bestcolor}{\textbf{79.08}} & \cellcolor{markcolor}\textcolor{bestcolor}{\textbf{78.31}} & \cellcolor{markcolor}\textcolor{bestcolor}{\textbf{83.44}} & \cellcolor{markcolor}\textbf{78.64} \\
 & $\Delta$ & \textcolor{applegreen}{+11.25} & \textcolor{applegreen}{+4.89} & \textcolor{applegreen}{+3.14} & \textcolor{applegreen}{+10.79} & \textcolor{applegreen}{+10.75} & \textcolor{applegreen}{+5.68} & \textcolor{applegreen}{+11.73} & \textcolor{applegreen}{+9.03} & \textcolor{applegreen}{+7.37} & \textcolor{applegreen}{+4.85} \\
\hline

\end{tabular}
\end{adjustbox}
\end{table}

\section{Concept Random Traverse}
\label{sec:random_order}

\subsection{Global Random Traverse}
\label{sec:global_random}

Global random traversal involves randomly traversing nodes throughout the AMR graph to capture concepts in a non-DFS manner, covering all sentences in the supporting documents. Concepts obtained in this setting have no specific positional relationship. The results are presented in Table~\ref{tab:aca_global_PopQA} and Table~\ref{tab:aca_global_EQQA}. These results reveal changes in $Acc.$ due to the concepts derived from global random traversal. Apart from a few $Acc.$ values higher than the concepts derived from the AMR-based concept distillation algorithm, most of the other $Acc.$ values corresponding to the same settings decrease. However, these changes in trend are not significant. We speculate that the reason is that global random traverse alters the positions of key concepts supporting the question within the concept set, leading to changes in LLMs' performance. This observation aligns with conclusions drawn by~\citet{liu2024lost}, which also demonstrates the necessity and rationality of using the DFS method to traverse the AMR graph to maintain the inherent positional relationship of concepts between the source texts.

\begin{table}[htbp]
\centering
\caption{The $Acc. \uparrow$ results on the PopQA dataset by concept obtained with the global random traverse. Any changes compared to the original proposed AMR-based concept distillation algorithm are \textcolor{downred}{marked}.}
\label{tab:aca_global_PopQA}
\begin{adjustbox}{width=1\linewidth}
\begin{tabular}{c|*{10}{c}}
\hline
\diagbox{LLMs}{$\mathcal{K}$} & 1 & 2 & 3 & 4 & 5 & 6 & 7 & 8 & 9 & 10 \\
\hline
{GPT-N-1.3B} & 34.96 & \textcolor{downred}{55.62} & \textcolor{downred}{60.43} & \textcolor{downred}{64.50} & \textcolor{downred}{63.35} & 76.82 & \textcolor{downred}{75.00} & 75.93 & 80.30 & 94.29 \\ \hline
{GPT-N-2.7B} & 43.09 & \textcolor{downred}{61.36} & \textcolor{downred}{63.51} & \textcolor{downred}{69.85} & 74.53 & 72.84 & 77.78 & 89.87 & 74.24 & 82.86 \\
\hline
{OPT-1.3B} & 47.56 & 63.12 & 63.51 & \textcolor{downred}{76.34} & \textcolor{downred}{72.05} & \textcolor{downred}{74.17} & 77.78 & 87.34 & 77.27 & 95.71 \\
\hline
{OPT-2.7B} & 49.86 & 65.11 & 68.96 & 71.76 &\textcolor{downred}{70.80} & \textcolor{downred}{80.79} & \textcolor{downred}{81.48} & 89.87 & 81.81 & \textcolor{downred}{90.00} \\
\hline
{Bloom-560m} & 35.91 & \textcolor{downred}{48.58} & \textcolor{downred}{53.08} & \textcolor{downred}{67.56} & 68.32 & 71.52 & 67.59 & 83.54 & 80.30 & 94.29 \\
\hline
{Bloom-1b1} & \textcolor{downred}{42.28} & 58.68 & \textcolor{downred}{65.40} & 71.76 & \textcolor{downred}{78.81} & 83.44 & 83.33 & \textcolor{downred}{88.61} & 86.36 & \textcolor{downred}{90.00} \\ 
\hline
{Bloom-1b7} & 47.43 & \textcolor{downred}{61.51} & 66.82 & \textcolor{downred}{74.81} & \textcolor{downred}{73.91} & 82.78 & 84.26 & \textcolor{downred}{93.67} & \textcolor{downred}{84.85} & \textcolor{downred}{95.71} \\
\hline
{Bloom-3b} & 49.73 & 60.67 & \textcolor{downred}{64.22} & \textcolor{downred}{71.76} & 74.53 & 79.47 & \textcolor{downred}{85.19} & 84.81 & \textcolor{downred}{83.33} & \textcolor{downred}{95.71} \\
\hline
{\footnotesize{LLaMA-2-chat-7b}} & \textcolor{downred}{62.73} & 74.06 & 74.64 & 76.72 & 78.26 & 87.42 & \textcolor{downred}{85.19} & 86.08 & \textcolor{downred}{86.36} & \textcolor{downred}{92.86} \\
\hline
{\footnotesize{LLaMA-2-chat-13b}} & 62.06 & 73.14 & 76.54 & 78.63 & 75.78 & 84.11 & \textcolor{downred}{82.41} & \textcolor{downred}{83.54} & \textcolor{downred}{78.79} & \textcolor{downred}{90.00} \\
\hline
\end{tabular}
\end{adjustbox}
\end{table}

\begin{table}[htbp]
\centering
\caption{The $Acc. \uparrow$ results on the EntityQuestions dataset by concept obtained with the global random traverse. The symbol definitions are the same with Table~\ref{tab:aca_global_PopQA}.}
\label{tab:aca_global_EQQA}
\begin{adjustbox}{width=1\linewidth}
\begin{tabular}{c|*{10}{c}}
\hline
\diagbox{LLMs}{$\mathcal{K}$} & 1 & 2 & 3 & 4 & 5 & 6 & 7 & 8 & 9 & 10 \\
\hline
{GPT-N-1.3B} & 36.74 & 51.38 & 57.61 & 59.91 & \textcolor{downred}{61.49} & 67.05 & 65.31 & \textcolor{downred}{57.83} & 71.17 &  \textcolor{downred}{66.99} \\
\hline
{GPT-N-2.7B} & 48.95 & 58.03 & 64.33 & \textcolor{downred}{66.96} & \textcolor{downred}{65.67} & \textcolor{downred}{70.83} & 72.96 & \textcolor{downred}{70.48} & 78.53 & \textcolor{downred}{73.79} \\
\hline
{OPT-1.3B} & 49.13 & 61.93 & 65.52 & \textcolor{downred}{67.62} & \textcolor{downred}{65.07} & \textcolor{downred}{71.97} & 70.41 & 65.66 & 75.46 & \textcolor{downred}{66.99} \\
\hline
{OPT-2.7B} & 55.36 & 66.99 & 68.81 & 72.91 & \textcolor{downred}{69.85} & \textcolor{downred}{77.65} & 67.86 & \textcolor{downred}{73.49} & 81.60 & \textcolor{downred}{66.99} \\
\hline
{Bloom-560m} & 32.91 & 44.90 & 48.36 & 47.14 & 49.85 & \textcolor{downred}{60.23} & 57.65 & \textcolor{downred}{51.81} & 66.87 & \textcolor{downred}{59.22} \\
\hline
{Bloom-1b1} & 43.87 & 58.74 & 63.88 & 66.96 & 66.27 & \textcolor{downred}{73.11} & 68.37 & \textcolor{downred}{62.65} & 71.17 & \textcolor{downred}{66.99} \\
\hline
{Bloom-1b7} & 50.87 & 62.29 & 63.58 & 67.84 & \textcolor{downred}{66.87} & 75.38 & \textcolor{downred}{68.37} & 73.49 & \textcolor{downred}{71.78} & \textcolor{downred}{74.76} \\
\hline
{Bloom-3b} & 49.97 & 58.65 & 64.03 & 65.86 & \textcolor{downred}{68.36} & \textcolor{downred}{74.62} & \textcolor{downred}{69.90} & \textcolor{downred}{72.89} & 74.23 & \textcolor{downred}{74.76} \\
\hline
{\footnotesize{LLaMA-2-chat-7b}} & 60.80 & 70.45 & 71.94 & 75.33 & 74.03 & 77.27 & 72.45 & 75.30 & 79.14 & \textcolor{downred}{73.79} \\
\hline
{\footnotesize{LLaMA-2-chat-13b}} & 63.61 & 74.37 & 74.18 & 81.72 & 80.00 & 81.44 & 79.08 & 78.31 & 83.44 & 78.64 \\
\hline
\end{tabular}
\end{adjustbox}
\end{table}

\subsection{Local Random Traverse}
\label{sec:local_random}

Local random traversal involves traversing nodes within a specific sentence of the AMR graph in a non-DFS manner to capture concepts. The concepts obtained through this approach do not maintain the positional relationship within individual sentences but preserve it between sentences in the retrieved supporting document. The results are shown in Table~\ref{tab:aca_local_PopQA} and Table~\ref{tab:aca_local_EQQA}. Similar to global random traversal, in this case, $Acc.$ shows a decreasing trend, but the magnitude of the decline is not substantial. We speculate that the reason is that, compared to the Vanilla methods, the word-level length of the distilled concept set is shorter, which reduces the sensitivity of LLMs in RAG to the position of key information when dealing with long contexts. The compression ratio of the original supporting documents using different methods will be elaborated in the next section.

\begin{table}[htbp]
\centering
\caption{The $Acc. \uparrow$ results on the PopQA dataset by concept obtained with the local random traverse. Any changes compared to the original proposed AMR-based concept distillation algorithm are \textcolor{downred}{marked}.}
\label{tab:aca_local_PopQA}
\begin{adjustbox}{width=1\linewidth}
\begin{tabular}{c|*{10}{c}}
\hline
\diagbox{LLMs}{$\mathcal{K}$} & 1 & 2 & 3 & 4 & 5 & 6 & 7 & 8 & 9 & 10 \\
\hline
{GPT-N-1.3B} &  \textcolor{downred}{34.55} &  \textcolor{downred}{55.39} &  \textcolor{downred}{60.90} & 64.89 &  \textcolor{downred}{63.98} & 76.82 & 75.93 & 75.95 & 80.30 & 94.29 \\ \hline
{GPT-N-2.7B} &  \textcolor{downred}{42.82} &  \textcolor{downred}{61.36} &  \textcolor{downred}{63.51} &  \textcolor{downred}{69.85} &  \textcolor{downred}{73.29} &  \textcolor{downred}{72.85} & 77.78 & 89.87 &  \textcolor{downred}{72.73} &  \textcolor{downred}{81.43} \\
\hline
{OPT-1.3B} & 47.56 &  \textcolor{downred}{62.82} &  \textcolor{downred}{63.03} &  \textcolor{downred}{75.57} & 72.67 &  \textcolor{downred}{75.50} &  \textcolor{downred}{76.85} & 87.34 & 77.27 & 95.71 \\
\hline
{OPT-2.7B} & 49.86 & 65.11 & 68.96 &  \textcolor{downred}{71.37} &  \textcolor{downred}{70.81} &  \textcolor{downred}{80.79} &  \textcolor{downred}{81.48} &  \textcolor{downred}{89.87} & 81.81 &  \textcolor{downred}{90.00} \\
\hline
{Bloom-560m} &  \textcolor{downred}{35.37} &  \textcolor{downred}{48.43} &  \textcolor{downred}{54.98} &  \textcolor{downred}{67.56} & 68.32 & 71.52 &  \textcolor{downred}{65.79} & 83.54 & 80.30 &  \textcolor{downred}{92.86} \\
\hline
{Bloom-1b1} & 42.55 &  \textcolor{downred}{58.45} &  \textcolor{downred}{65.88} &  \textcolor{downred}{71.37} & 79.50 & 83.44 & 83.33 &  \textcolor{downred}{87.34} & 86.36 &  \textcolor{downred}{90.00} \\ 
\hline
{Bloom-1b7} &  \textcolor{downred}{47.56} &  \textcolor{downred}{61.44} & 66.82 & 75.19 &  \textcolor{downred}{73.91} & 82.78 &  \textcolor{downred}{83.33} &  \textcolor{downred}{93.67} &  \textcolor{downred}{84.85} &  \textcolor{downred}{94.29} \\
\hline
{Bloom-3b} &  \textcolor{downred}{49.59} & 60.67 &  \textcolor{downred}{64.21} &  \textcolor{downred}{71.76} & 74.53 & 79.47 & 86.11 & 84.81 &  \textcolor{downred}{83.33} &  \textcolor{downred}{94.29} \\
\hline
{\footnotesize{LLaMA-2-chat-7b}} &  \textcolor{downred}{62.60} & 74.06 & 74.64 & 76.72 &  \textcolor{downred}{78.88} & 87.42 &  \textcolor{downred}{83.33} & 86.08 &  \textcolor{downred}{84.85} &  \textcolor{downred}{91.43} \\
\hline
{\footnotesize{LLaMA-2-chat-13b}} &  \textcolor{downred}{61.52} &  \textcolor{downred}{73.60} &  \textcolor{downred}{76.30} &  \textcolor{downred}{77.86} &  \textcolor{downred}{75.16} & 84.11 &  \textcolor{downred}{83.33} &  \textcolor{downred}{83.54} &  \textcolor{downred}{75.76} &  \textcolor{downred}{91.43} \\
\hline
\end{tabular}
\end{adjustbox}
\end{table}

\begin{table}[htbp]
\centering
\caption{The $Acc. \uparrow$ results on the EntityQuestions dataset by concept obtained with the local random traverse. The symbol definitions are the same with Table~\ref{tab:aca_local_PopQA}.}
\label{tab:aca_local_EQQA}
\begin{adjustbox}{width=1\linewidth}
\begin{tabular}{c|*{10}{c}}
\hline
\diagbox{LLMs}{$\mathcal{K}$} & 1 & 2 & 3 & 4 & 5 & 6 & 7 & 8 & 9 & 10 \\
\hline
{GPT-N-1.3B} & \textcolor{downred}{36.44} & \textcolor{downred}{51.20} & 57.61 & 59.91 & \textcolor{downred}{61.49} & \textcolor{downred}{66.67} & 65.31 & 62.65 & 71.17 & \textcolor{downred}{66.02} \\
\hline
{GPT-N-2.7B} & \textcolor{downred}{48.89} & \textcolor{downred}{58.30} & \textcolor{downred}{64.48} & 67.00 & \textcolor{downred}{65.67} & \textcolor{downred}{72.21} & 72.96 & 71.08 & 78.53 & 79.61 \\
\hline
{OPT-1.3B} & \textcolor{downred}{49.07} & \textcolor{downred}{62.20} & \textcolor{downred}{65.22} & 67.62 & 68.06 & 74.24 & 70.41 & 65.66 & \textcolor{downred}{73.62} & 72.82 \\
\hline
{OPT-2.7B} & 55.36 & \textcolor{downred}{66.90} & 68.81 & \textcolor{downred}{72.69} & 72.83 & \textcolor{downred}{77.65} & \textcolor{downred}{67.35} & \textcolor{downred}{73.49} & \textcolor{downred}{82.20} & 76.70 \\
\hline
{Bloom-560m} & 32.91 & 44.90 & \textcolor{downred}{47.76} & 47.14 & \textcolor{downred}{49.25} & 65.53 & \textcolor{downred}{56.63} & 54.22 & \textcolor{downred}{65.03} & \textcolor{downred}{60.19} \\
\hline
{Bloom-1b1} & 43.87 & \textcolor{downred}{58.65} & 63.88 & 66.96 & \textcolor{downred}{65.67} & \textcolor{downred}{72.73} & 68.37 & 71.08 & \textcolor{downred}{70.55} & 77.67 \\
\hline
{Bloom-1b7} & 50.87 & 62.29 & \textcolor{downred}{64.03} & 67.84 & 72.24 & 75.38 & \textcolor{downred}{68.37} & \textcolor{downred}{74.09} & \textcolor{downred}{71.16} & \textcolor{downred}{73.79} \\
\hline
{Bloom-3b} & \textcolor{downred}{49.85} & \textcolor{downred}{58.39} & 64.03 & 65.86 & \textcolor{downred}{68.36} & 75.38 & \textcolor{downred}{69.90} & 77.11 & \textcolor{downred}{75.46} & 78.64 \\
\hline
{\footnotesize{LLaMA-2-chat-7b}} & \textcolor{downred}{59.90} & \textcolor{downred}{70.90} & \textcolor{downred}{72.39} & \textcolor{downred}{73.73} & 74.03 & 77.27 & 72.45 & 75.30 & \textcolor{downred}{80.37} & \textcolor{downred}{74.76} \\
\hline
{\footnotesize{LLaMA-2-chat-13b}} & \textcolor{downred}{64.09} & \textcolor{downred}{73.82} & \textcolor{downred}{73.43} & 81.72 & 80.00 & \textcolor{downred}{80.30} & \textcolor{downred}{78.57} & \textcolor{downred}{77.11} & \textcolor{downred}{82.82} & 78.64 \\
\hline
\end{tabular}
\end{adjustbox}
\end{table}

\section{Compression Ratio}
\label{sec:compression_ratio}

\begin{figure*}[htbp]
 \centering
 \includegraphics[width=1\textwidth]{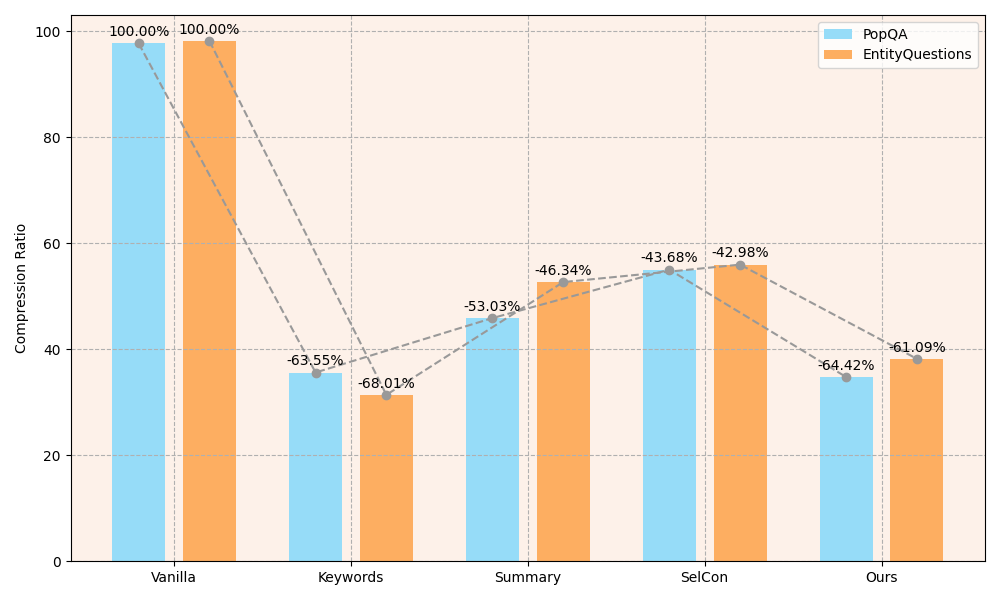}
 \caption{The word-level compression ratio of different context compression methods, while we take the length of the "Vanilla" method as the standard (100.00\%), and the numbers above the histogram represent the length reduction compared to "Vanilla".}
 \label{fig:compression_ratio}
\end{figure*}

This section compares the word-level compression ratio achieved by different context compression methods with the results shown in Fig.~\ref{fig:compression_ratio}. The findings demonstrate that the proposed AMR-based context distillation algorithm achieves competitive context-compression performance. Compared to the standard Vanilla method, our method can compress the average word-level length by over 60\% while preserving the core concepts to the greatest extent possible. Notably, our method achieves the highest compression ratio on the PopQA dataset. In contrast, although the Keywords-based compression method exhibits a 6.92\% higher compression ratio on the EntityQuestions dataset compared to ours, its performance lags behind the proposed algorithm in preserving core concepts. Compared to the other two baselines, our method shows a significant advantage in terms of compression ratio.

\section{Inference Latency}
\label{sec:inference_latency}

\begin{figure*}[htbp]
 \centering
 \includegraphics[width=1\textwidth]{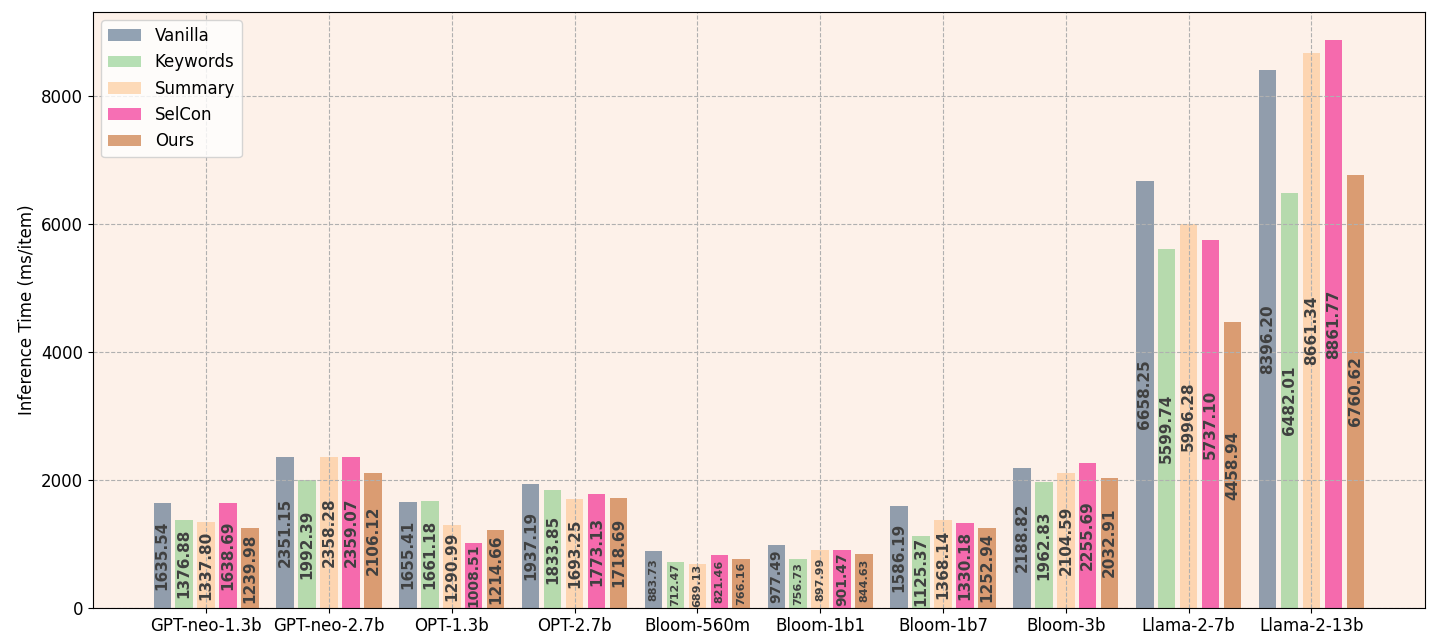}
 \caption{The inference time comparison of different context compression methods. The values in the histogram represent the inference time for each model on a single item, measured in milliseconds.}
 \label{fig:time}
\end{figure*}

This section analyses the inference latency by providing the inference time comparison of backbone LLMs with different compression methods, the result is shown in Fig.~\ref{fig:time}. The findings reveal that the context compressed by our method has a significant advantage in the inference of most LLMs. Moreover, the context compressed by the Keywords method also exhibits shorter inference times, which aligns with the trend observed in the previous compression ratio analysis.

\end{document}